# The power of text similarity in identifying AI-LLM paraphrased documents: The case of BBC news articles and ChatGPT

**Konstantinos F. Xylogiannopoulos [a,b], Petros Xanthopoulos [c], Panagiotis Karampelas [d], Georgios Bakamitsos[e]**

[a] Stetson University, School of Business Administration, DeLand, FL, USA, kxylogiannopoulos@stetson.edu

[b] University of Calgary, Calgary, AB, Canada, kostasfx@yahoo.gr

[c] Stetson University, Business Systems & Analytics Department, School of Business Administration, DeLand, FL, USA, pxanthopoulos@stetson.edu

[d] Hellenic Air Force Academy, Dekeleia, Greece, panagiotis.karampelas@hafa.haf.gr

[e] Stetson University, Marketing Department, School of Business Administration, DeLand, FL, USA, bakamitsos@stetson.edu

**Abstract**

Generative AI paraphrased text can be used for copyright infringement and the AI paraphrased content can deprive substantial revenue from original content creators. Despite this recent surge of malicious use of generative AI, there are few academic publications that research this threat. In this article, we demonstrate the ability of pattern-based similarity detection for AI paraphrased news recognition. We propose an algorithmic scheme, which is not limited to detect whether an article is an AI paraphrase, but, more importantly, to identify that the source of infringement is the ChatGPT. The proposed method is tested with a benchmark dataset specifically created for this task that incorporates real articles from BBC, incorporating a total of 2,224 articles across five different news categories, as well as 2,224 paraphrased articles created with ChatGPT. Results show that our pattern similarity-based method, that makes no use of deep learning, can detect ChatGPT assisted paraphrased articles at percentages 96.23% for accuracy, 96.25% for precision, 96.21% for sensitivity, 96.25% for specificity and 96.23% for F1 score.

## 1. Introduction

A paraphrase is defined as texts or phrases that convey the same meaning by using different wording (Bhagat & Hovy, 2013). Since 2022 with the release of the first public version of ChatGPT by OpenAI (Wu et al., 2023) and the subsequent release of other Large Language Models (LLMs) Artificial Intelligence (AI) tools, text paraphrase applications were able to scale a lot faster than before. Text paraphrase has several useful applications. Amazon in 2023 released an AI assisted review summarization service to improve customer experience (Schermerhorn, 2023). Currently, several online retailers use text paraphrase for summarizing user reviews and providing a short overview to users who seek to make an informed decision about a service or a product. LegalZoom, an online provider of legal related business services, provides AI-based commercial legal document summarization tools that allow their users to faster review and gain insights of legal documents (LegalZoom, 2024). Several other providers such as Rephrase, Linguix, Word AI, CoderDuck's and Study Crumb's offer commercial paraphrasing services for improving engagement, search engine optimization (SEO) friendliness and marketing engagement of text content. In that respect, there is a vested interest in the development of better text paraphrase tools and an application industry that is built around these new technical capabilities. In a recent report, McKinsey has identified that Generative AI will be able to add trillions of dollars to the global GDP with 75% of this new value being associated with customer operations, marketing and sales, software engineering and R&D (Chui et al., 2023).

On the other side, AI assisted paraphrase has several malicious applications. In our previous work, we argued that generative AI tools can be used to artificially alter the online reputation of products or services (K. F. Xylogiannopoulos et al., 2024a).Furthermore, a lot of book authors have witnessed loss of sale revenue mainly attributed to AI imitation self-published books that are competing with the original on book selling platforms such as Amazon (Knibbs, 2024). New book releases are more prone to such malicious attacks and organizations such as AuthorsGuild, America's oldest professional organization of writers, is lobbying to impose regulations to labels AI-generated content on such marketplaces (AuthorsGuild, 2024). Paraphrased sham content also affects other areas such as online journalism. According to a recent investigation by Newsguard, a misinformation tracking platform, it was found that 37 websites used AI assisted paraphrase to repost news articles from reputable sources such as CNN, New York Times and Reuters as theirs without proper attribution of the content source. In fact, such platforms are known as copycat content farms, use AI- assisted plagiarism to inflate the volume of their content and therefore appear higher in search engine results. Furthermore, popular plagiarism detector tools, such as Grammarly, failed to detect the paraphrased news articles at a rate of 79% (Brewster et al., 2023).

The inability of known news content aggregation services such as Google news to detect and ban such plagiarized content has also been noted (Weintraub, 2023) although sham content usually appears in content searches only when one searches for articles for the last 24 hours (Barnes, 2024).

In a recent research opinion poll of 1600 researchers, the 68% expressed the thought that AI will make plagiarism easier to construct and harder to detect (Van Noorden & Perkel, 2023). Despite this concern of this new-era type of plagiarism the academic research hasn't been able to catch up with this challenge. In fact, recent studies have demonstrated that existing AI-language detection platforms fall short in detection attacks that involved plagiarism. Although approaches such as semantic similarity search have been proposed to circumvent this limitation (Krishna et al., 2023), a recent study has demonstrated that existing approaches have limited scalability and with the appropriate paraphrase mechanism AI-assisted similarity is virtually undetected under the current existing commercial mechanisms (Sadasivan et al., 2024). Recently, we demonstrated the existence of a consistence word similarity pattern between original text and their AI-assisted paraphrase counterparts (K. F. Xylogiannopoulos et al., 2024a) and in subsequent work we demonstrate the ability of this characteristic to be used as a basis for an AI paraphrased fake review detection algorithm that was tested with original and fake reviews from TripAdvisor (K. F. Xylogiannopoulos et al., 2024b).

### 1.1. Contribution/novelty

This paper is the natural extension of our previous work to the field of AI-assisted paraphrased news article detection problem. The core of the contribution and novelty of the present work can be summarized in the following bullet points:

- Novel dataset that is based on real articles published on BBC, which is old enough to guarantee that it is AI-polluted free, and their paraphrases created by ChatGPT 4.0. To the best of our knowledge there is no other such benchmark dataset in the literature.

- This work adds to the limited literature of AI assisted paraphrase detection. Most literature deals with the AI text detection problem which differs substantially from the problem that is presented here.

- The current research introduces algorithmic text similarity detection as a method for detecting paraphrased news articles. To the best of our knowledge, there has not been another paper incorporating such approach.

- The proposed, algorithmic, methodology departs methodologically from the core body of knowledge detection/classification approaches that are based on models from machine learning and more specifically deep learning. We hope that this work will open a whole new direction to the field of AI paraphrase detection with scalable, cost effective and computationally efficient pattern detection algorithms.

The rest of the paper is structured as follows: In section 2, we provide a literature review with methods and techniques associated with news articles paraphrase detection. In section 3, we describe our materials and methods that include the novel benchmark dataset as well as the proposed text-based similarity detection method. In section 4, we provide the results. In section 5. we discuss the results and provide benefits and limitations of the proposed methodology. Last in section 6, we conclude the presentation of our work.

### 1.2. Research objectives

The overarching goal of our research is to study the usefulness of word similarity-based patterns in distinguishing human from ChatGPT paraphrased text. In this paper we aim to achieve the following:

- To determine whether it is possible to derive a similarity based feature that is statistically significant when comparing human and AI-assisted paraphrased text.
- To determine whether this feature can be incorporated into a classifier able to correctly classify texts generated from ChatGPT and other sources.
- To achieve the first two objectives without defaulting to traditional deep learning approaches that require vast amounts of data to create their models, advanced and extended computational resources to be executed, and significant cost to operate, update and maintain.

In summary, we propose an algorithmic based methodology that uses word pattern similarities to distinguish between original human news articles and their corresponding ChatGPT paraphrases, without any modelling or previously acquired knowledge.

## 2. Literature review

### 2.1. Paraphrase creation

#### 2.1.1. Pre LLM-era paraphrase

Paraphrase detection as a tool has existed long before the advent of LLM based models. Most of the early work used the Microsoft paraphrase corpus (MSRP) as benchmark (Dolan & Brockett, 2005). It consists of 5,800 pairs of sentences extracted from the web along with human annotations as to whether the pairs are paraphrases or not. Subsequent datasets included TUC (Xu et al., 2015), ParaNMT-50M (Wieting & Gimpel, 2018), MSCOCO (Lin et al., 2014), Quora (Fu et al., 2019) and ParaSCI (Dong et al., 2021). The common characteristic of these datasets is that the average length and character count of the pairs are limited between 10-22 words and 50-120 characters respectively. Along with English corpora, there have been corpora in other languages such as Russian (Pivovarova et al., 2018), Turkish (Eyecioglu & Keller, 2018),

Arabic (Alian et al., 2021) and Armenian (Malajyan et al., 2020). The PPDB dataset includes over a total of one billion sentences that covers more than 20 languages (Ganitkevitch & Callison-Burch, 2014). The limited word count size of the individual samples in these datasets have dictated the solution approach algorithms with the two most prominent literature approaches being semantic similarity combined with deep learning (Agarwal et al., 2018; AL-Smadi et al., 2017; Shakeel et al., 2020; Vrbanec & Meštrović, 2020). However, it has been pointed out that approaches that have succeeded in a certain dataset don't generalize well when applied to the other (Agarwal et al., 2018). This is indicative of failure to avoid overfitting.

### 2.1.2. Paraphrase in the era of LLM's

Since the launch of the original release of ChatGPT in 2022 and its subsequent revisions (ChatGPT 4.0 in this case) along with other large language generative AI models such as Google Gemini and Llama by Meta, generating paraphrases of relatively long text is a capability that can scale up the beneficial applications along with the malicious use of these tools. Some researchers have used these new tools in order to easily expand the existing corpora (Wahle et al., 2023) and other works by creating paragraph size paraphrases of original text excerpts (Vrbanec & Meštrović, 2023; Wahle et al., 2022). Many researchers have pointed out that the problem of detecting AI-created paraphrased text is a significant more complex task compared to prior ones. Krishna et al., (2023) demonstrated the inability of commercial AI detection tools to detect AI generated paraphrases and at the same time suggested semantic similarity to circumvent this shortcoming. However, Sadasivan et al., (2024) pointed out a practical suggesting that there is no reliable method for detecting paraphrased text generated by LLM models. The ability of current AI detection tools to detect paraphrased AI generated text was further challenged by a number of other more recent studies (Chakraborty et al., 2023; Huang et al., 2024; Steponenaite & Barakat, 2023).

### 2.1.3. AI Paraphrase detection

So far in the literature there have been a handful of approaches that test AI detection methods to AI paraphrased text instances. These approaches include adversarial learning (Hu et al., 2023), cross-entropy (Li et al., 2024), Multi-level Detection Fine-grained (MFD) approach (Tao et al., 2024), and metric based approach (Le & Tran, 2025). Becker et al., (2023) underlined the importance for more machine generated benchmark datasets. In (K. F. Xylogiannopoulos et al., 2024a) we observed a consistent text similarity pattern that appears between original and AI-paraphrased hotel reviews from TripAdvisor and in (K. F. Xylogiannopoulos et al., 2024b) we established an efficient paraphrased review detection for the same dataset.

## 3. Materials and Methods

### 3.1. Dataset

In this study, a dataset of news articles was used, sourced from BBC News between 2004 and 2005, comprising 2,224 articles categorized into five topics (Greene & Cunningham, 2006). Specifically, the dataset includes 509 articles in Business, 386 in Entertainment, 417 in Politics, 511 in Sports, and 401 in Technology. Table 1 provides a summary of the dataset's statistics, detailing the number of articles per category, as well as the maximum, minimum, mean, and standard deviation of word counts per article. These statistics are particularly relevant, as article length is a key limitation affecting ChatGPT's output, especially for outliers of 800 words and above (Fig. 1), potentially influencing the results, as discussed later.

It is also important to point out that during our research for an appropriate dataset, it was not possible to discover and acquire any other one that could meet the minimum attributes required for such an important and difficult task, i.e., (a) be solely based on news organization articles, (b) be freely redistributable without copyright restrictions, (c) be inspected and validated by scientists for its integrity and quality, (d) be large and diverse enough to allow proper evaluation, (e) have balanced articles word length distribution, i.e., not solely very short or very long, and (f) be old enough to ensure that it does not contain, accidentaly or intentionally, any LLM produced text.

*Table 1* Dataset descriptive statistics

| Articles Category | Articles count | Max Words | Min Words | Mean Words | Standard Deviation Words |
|---|---|---|---|---|---|
| Business | 509 | 909 | 138 | 333.10 | 137.87 |
| Entertainment | 386 | 3554 | 138 | 334.37 | 267.65 |
| Politics | 417 | 4487 | 86 | 456.28 | 305.28 |
| Sports | 511 | 1724 | 114 | 336.12 | 193.93 |
| Technology | 401 | 2991 | 158 | 507.79 | 243.08 |

In Fig. 1 more advanced statistics are presented in a graphical form and, additionally, the figure allows to observe distribution and significant outliers.

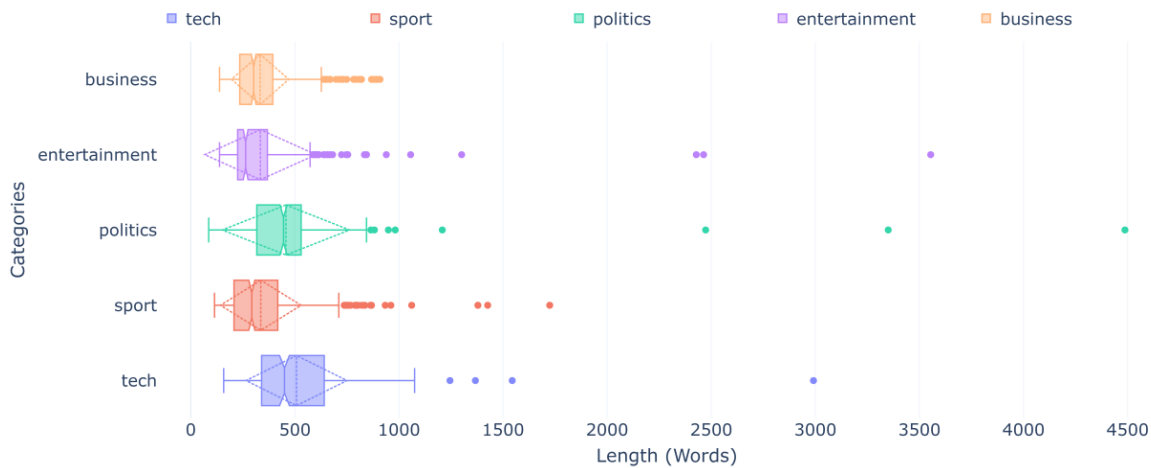

*Fig. 1 Descriptive statistics of article word count for the original BBC article dataset for the five respective news categories.*

For each BBC article, we created a paraphrase by using ChatGPT. For ChatGPT, we used the prompt "Please paraphrase the following text: <BBC article to paraphrase>". The implementation was done through a python script that used OpenAI's Application Programming Interface (API). We used a temperature parameter equal to 1. Temperature controls the variability in LLMs response. Temperature closer to 0 indicates more response determinism and closer to 1 more variability (randomness). Therefore, for each of the 2,224 original articles we created a ChatGPT paraphrase article yielding a total of 2,224 Original + 2,224 ChatGPT = 4,448 total articles.

### 3.2. Methodology

Before we introduce our methodology, we find it useful to introduce some definitions that will be used throughout the rest of the paper.

**Definition 1**: We define as "Original*"* a text produced by a news organization that might have been suffered plagiarism by ChatGPT. For the purpose of this work, original texts are only these included in the BBC dataset as described in (Greene & Cunningham, 2006).

**Definition 2**: We define as "Suspicious" a paraphrased text of an Original text, which has been produced by ChatGPT (with temperature 1), and is the text under investigation. Suspicious texts are used for evaluating our methodology. Our detection scheme has no knowledge of a suspicious text identity.

**Definition 3**: We define as "Reference" a paraphrased text of an Original text, which has been produced by ChatGPT (with temperature 0), and is produced as part of our detection methodology for similarity quantification purposes against its Original and all Suspicious texts.

The cornerstone of our proposed methodology is based on a key observation that was introduced in previous work according to which an Original text has relatively low similarity compared to a Reference and Suspicious text, but similarity between Reference and Suspicious texts is substantially higher. This was the case for various temperature values tested (K. F. Xylogiannopoulos et al., 2024a). In this framework, similarity has been found to be useful in distinguishing between real and paraphrased TripAdvisor reviews (K. F. Xylogiannopoulos et al., 2024b).

For the case of the ChatGPT paraphrased news articles detection problem of this paper, we take advantage of the fact that a Suspicious article from an external source, which is a potential paraphrase of a news agency organization original article, it can be easily mapped to that original text. Therefore, the Suspicious article must be tested for potential plagiarism against a single, already identified, article and not the entire news agency organization database. For example, if a series of online articles, e.g., about a US President Office announcement on a specific policy, is identified as a potential counterfeit paraphrase of the original article of a news organization, it only needs to be compared in an one-to-one base against the specific original article. In the worst case, where more than one original article exist due to, e.g., continuous and narrow time spanned updates on the same topic, again a potential counterfeit paraphrase has to be examined and compared in an one-to-one base against every recent, topic specific, original article and not against the complete pool of previously published articles, which are irrelevant and outdated to the topic.

Our methodology can be divided into two phases, for simplicity reasons, in a waterfall architecture where the second phase is only executed if the first (simpler) phase fails to directly classify a text. In the first phase, we directly compare the Original article text to the Suspicious text for similarities (Fig. 2).

If the similarities percentage is very large, e.g., 80% then we can classify the Suspicious text as a paraphrase and stop the process, without investigating Suspicious text origin, since it is practically a copy-paste version of the Original article. However, as it has been already mentioned, this is not very common for ChatGPT, even for paraphrases with temperature zero (K. F. Xylogiannopoulos et al., 2024a). An example of the first phase is shown in Fig. 3 where the common patterns of length three or above are depicted on both texts, Original (first column) and Suspicious (second column). Indeed, there are some common patterns, yet the similarity ratio between the two texts is very low considering that the Suspicious text is a direct ChatGPT paraphrase of temperature one of the Original.

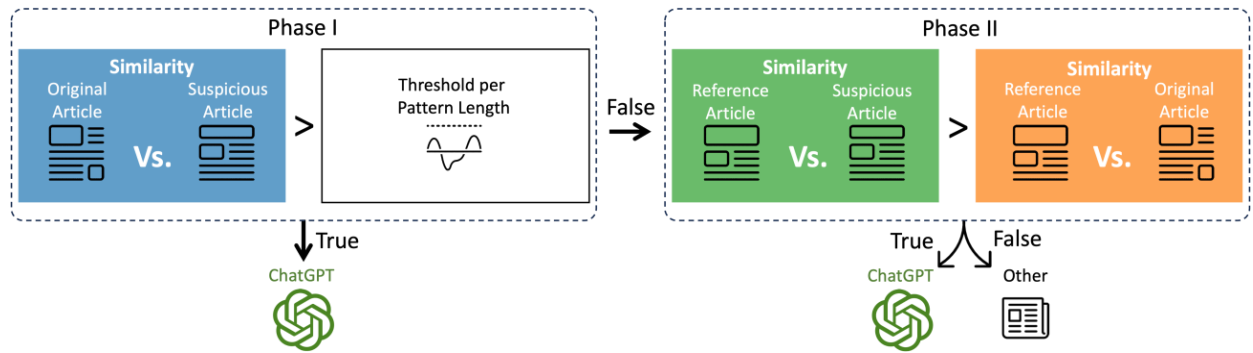

*Fig. 2 A high level overview of the two-phase proposed methodology*

Fig. 3 An example of an Original (left) and a Suspicious (right) text along with their marked similarities of three word or longer patterns

In the most common case where the above check fails, e.g., when temperature is one, the second, most important phase of the methodology is executed. Now, a ChatGPT paraphrase (Reference) of the Original article text is created and it is compared for similarities against both the Original article and the Suspicious text (Fig. 3). In this case, the newly created paraphrase is also expected to have very low similarity score with the Original article text. If the similarity between the text of the intentionally created paraphrase for testing purposes and the Suspicious text (REFERENCE-SUSPICIOUS) is higher than the similarity between the Reference and the Original (REFERENCE-ORIGINAL) then we classify the Suspicious text as ChatGPT paraphrase, otherwise not (Fig. 3). Additionally, the higher is the spread between the two similarity ratios, the more confident we are about the probability of the Suspicious text to be a ChatGPT generated paraphrase of the Original.

Expanding the example of Fig. 3, a ChatGPT paraphrase of temperature zero of the Original text has been created and compared to both Original and Suspicious texts (Fig. 4). In Fig. 4, we have the colored presentation of similar patterns between the Original text (first column) and the Reference, Suspicious text (second column) and Reference and both Original and Suspicious with Reference text (third column). Same background color has been used in the comparisons between REFERENCE-ORIGINAL and font color in the comparisons between REFERENCE-SUSPICIOUS, for easier visual identification of the patterns. What it is easy to observe in this case is the significantly higher similarity between Reference and Suspicious texts compared to Reference and Original. Especially, in the third column (Reference text), the green font text outperforms the orange background text revealing the drastically higher similarity between the Reference and the Suspicious text, allowing us to classify the Suspicious as a high probable ChatGPT paraphrase of the Original. This simple example lays the ground for our proposed methodology that is subsequently presented.

| Original - Reference | Suspicious - Reference | Reference - Original & Suspicious |
|---|---|---|
| Another three US insurance executives have pleaded guilty to fraud charges stemming from an ongoing investigation into industry malpractice. Two executives from American International Group (AIG) and one from Marsh & McLennan were the latest. The investigation by New York attorney general Eliot Spitzer has now obtained nine guilty pleas. The highest ranking executive pleading guilty on Tuesday was former Marsh senior vice president Joshua Bewlay. He admitted one felony count of scheming to defraud and faces up to four years in prison. A Marsh spokeswoman said Mr Bewlay was no longer with the company. Mr Spitzer's investigation of the US insurance industry looked at whether companies rigged bids and fixed prices. Last month Marsh agreed to pay $850m (£415m) to settle a lawsuit filed by Mr Spitzer, but under the settlement it "neither admits nor denies the allegations". | Three more executives in the US insurance sector have admitted to committing fraud, as a result of a widespread probe into unethical practices within the industry. This group includes two individuals from American International Group (AIG) and one from Marsh & McLennan. The inquiry, led by New York's Attorney General Eliot Spitzer, has now resulted in a total of nine confessions. On Tuesday, the most senior figure to confess was Joshua Bewlay, a former senior vice president of Marsh, who conceded to a felony charge of engaging in a fraudulent scheme and now potentially faces a prison sentence of up to four years. A representative from Marsh stated that Bewlay is no longer employed by the firm. Spitzer's examination into the American insurance industry scrutinizes accusations of bid-rigging and price-fixing. Marsh settled with Spitzer last month for $850 million to resolve a lawsuit, although the company did not acknowledge any wrongdoing as part of the agreement. | Three more executives from the US insurance sector have admitted to engaging in fraudulent activities as part of a broader probe into unethical practices within the industry. This includes two individuals from American International Group (AIG) and another from Marsh & McLennan. These admissions bring the total number of guilty pleas to nine in the investigation led by New York Attorney General Eliot Spitzer. The most senior of the executives to confess on Tuesday was Joshua Bewlay, a former senior vice president at Marsh, who acknowledged his involvement in a felony fraud scheme and now potentially faces a prison sentence of up to four years. A representative from Marsh has confirmed that Bewlay is no longer employed by the firm. Spitzer's scrutiny of the insurance industry in the US has been examining the possibility of bid-rigging and price-fixing. In the previous month, Marsh consented to pay $850 million to resolve a lawsuit initiated by Spitzer, although the company did not concede to any of the claims in the settlement. |

*Fig. 4 An example of an Original text (left) a Suspicious text (center) and a Reference text (right) along with their pairwise similar patterns shown in different colors*

Moreover, as it can be inferred by the above introduction of our methodology, the reason to have two distinct phases is mainly to control the cost of the methodology in cases where the initially compared texts are practically identical. Therefore, there is no need to create the Reference text and increase the overall cost of the methodology especially for a news agency organization to impose multiple investigations of its entire publications database.

### 3.2.1. Dataset Cleaning Stage

The first stage for both phases is the text cleaning. Special characters must be removed to guarantee as much as possible simplification of the text, since the analysis is pattern based. For this purpose, characters like: ".", ",", ";", "!", "?", ":", "(", ")", "-", "*", "[", "]", "{", "}", "\\", "/", "#" have been replaced by space in the text.

### 3.2.2. Pattern Detection Stage

The most fundamental part of our methodology is the common pattern detection among Original, Suspicious and Reference texts. Practically, we want to detect repeated patterns between all texts combined. This is achieved through the All Repeated Patterns Detection (ARPaD) algorithm and the Longest Expected Repeated Pattern Reduced Suffix Array (LERP-RSA) data structure. The ARPaD algorithm can detect every repeated pattern that exists in a dataset comprised of a single text or multiple texts, representing standard human language simple text to even non-discrete datasets such as real-value time series. The LERP-RSA data structure is a special case of suffix array data structure, specifically designed to allow efficient ARPaD execution. It has many special attributes, such as alphabet pre-classification, which permits parallel ARPaD execution on significantly smaller datasets or the creation of a combined data structure from multiple datasets on which ARPaD detects patterns that are common between them.

Both the algorithm and the data structure are computationally efficient, and this has been proven through massive datasets analyses, in combination with other methodologies for solving complex problems. Example applications of LERP-RSA and ARPaD in big data, which prove their ability to process and analyze large amounts of data, can be found in Number Theory with the detection of repeated patterns in the first one trillion ($10^{12}$) digits of π (K. Xylogiannopoulos, 2017), in Market Basket Analysis for the on-line analysis of 96 million simulated transactions dataset (K. Xylogiannopoulos et al., 2018), in Network Security and Data Analytics with real time analysis of approximately 12 million packets per second (K. F. Xylogiannopoulos et al., 2021) and in Bioinformatics and Computational Biology for the common biological patterns detection in 300 thousand complete genomes (K. F. Xylogiannopoulos, 2022).

The first step of the pattern detection stage is to create the appropriate LERP-RSA data structure that will be used by the ARPaD algorithm. For pattern detection of the specific use case texts presented here, the data structure is created with the ability to detect patterns from 3 up to 15 words. There is no need for longer patterns detection, because, according to the Probabilistic Existence of the Longest Expected Repeated

Pattern Theorem, longer repetitions are not expected (K. Xylogiannopoulos, 2017; K. Xylogiannopoulos et al., 2016).

In this study, first the LERP-RSA data structure is created and then ARPaD algorithm is executed to detect the repeated patterns that have length greater than or equal to three words. This threshold is chosen because patterns of one or two words are too trivial and cannot produce results that are meaningful or, even worse, can produce misleading results by introducing noise. Although both processes can be executed in parallel to achieve significant time execution performance, yet this is not necessary due to the dataset size.

### 3.2.3. Similarity Ratio Calculation Stage

The next step of similarity classification between texts is to split the patterns in classes based on which pair of texts they belong. We define as pairs every two text combination, i.e., ORIGINAL-SUSPICIOUS, REFERENCE-ORIGINAL and REFERENCE-SUSPICIOUS. Therefore, every class has patterns that exist in both texts, e.g., patterns in both Original and Suspicious.

Yet, we must note that the similarity ratio between two texts, e.g., ORIGINAL-SUSPICIOUS, can have two values depending on which text length is used as base, since compared texts rarely have the same length. For example, if the Original text has 100 words, Suspicious has 200 and they have 30 words in common, then the Original is 30% similar to the Suspicious while the Suspicious is 15% similar to the Original. Although this is not a drawback for the methodology, it can be addressed by normalizing the lengths or by requiring the Reference text to explicitly have approximate length to the one or the other.

### 3.2.4. Similarities Classification

The stages described so far are used to calculate similarity ratios between texts that they will be used to classify a text as ChatGPT or not. In the first phase of this process, the base similarity ratios are calculated with respect to the Original and Suspicious texts. The overall architecture of the method that incudes, data cleaning, pattern detection and similarity ratio quantification, can be seen in Fig. 5.

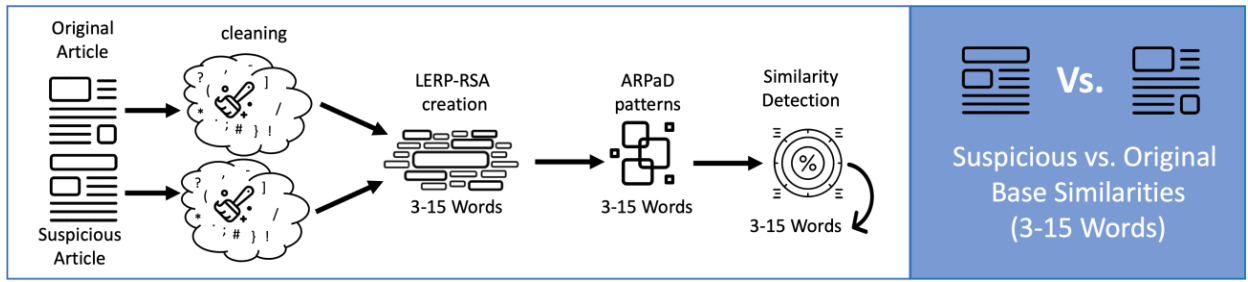

*Fig. 5 Phase I base similarities creation*

Then these similarities are compared for each pattern length against a pre-determined similarity pattern length threshold. The threshold can be determined empirically or by a high value, e.g., 80% similarity or above implies obvious paraphrasing. If the similarity is above the threshold, then the value one is recorded, multiplied by a weight, otherwise is recorded as minus one times the weight. Different weight values could potentially be used to emphasize patterns of certain length, which may be considered more important, e.g., longer patterns compared to shorter (Fig. 6). After completing the process for all pattern lengths then the results are added. If the sum is positive, hence more similarities exist above the threshold per pattern length (and multiplied by weight 1), then the Suspicious text is classified as ChatGPT paraphrase or, otherwise, it cannot be classified in this phase and we need to proceed to the second one (Fig. 6). One way to determine the threshold is to analyze synthetically created suspicious texts created by ChatGPT and comparing their similarities to the corresponding Original text.

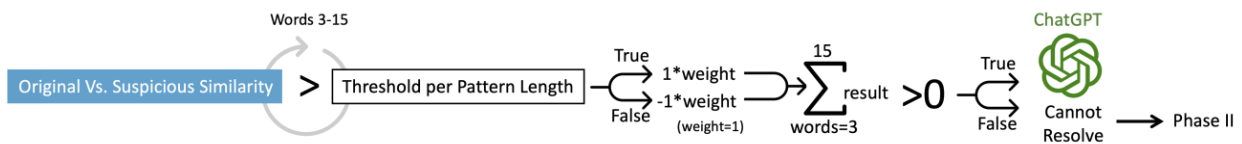

*Fig. 6 Phase I classification determination*

For the second phase, two similarities per pattern length must be determined, one for the REFERENCE-ORIGINAL and one for the REFERENCE-SUSPICIOUS (Fig. 7). This leads to two groups of similarities for each pair of texts.

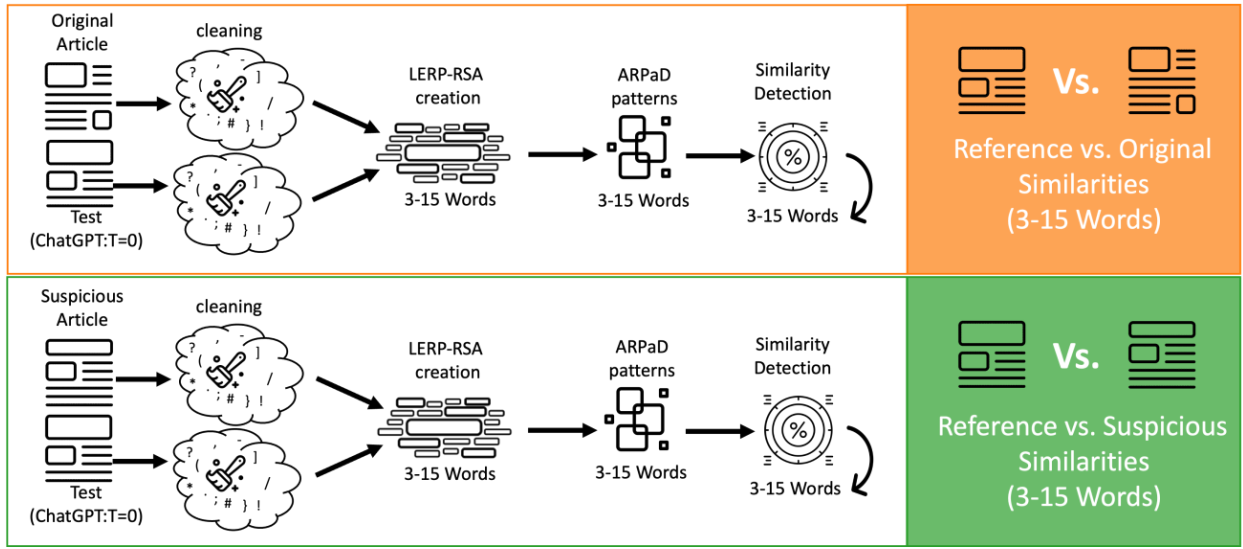

*Fig. 7 Phase II similarities creation*

The main difference from phase one is that now we compare the two groups of similarities per pattern length against each other and not against a threshold. Therefore, if the similarity between REFERENCE-SUSPICIOUS is greater than the REFERENCE-ORIGINAL then value one is recorded, multiplied by a pattern length specific weight, as defined previously (Fig. 8). Completing the checks, the results are added and if the output is positive then the Suspicious text is classified as ChatGPT paraphrase, otherwise as other source.

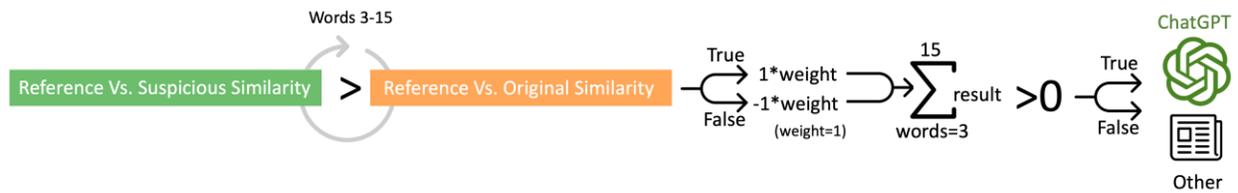

*Fig. 8 Phase II classification determination*

## 4. Results

### 4.1. Methodology Verification

To verify our methodology, we used the BBC news dataset described in Section 4.1. The dataset includes articles that fall under five categories namely business, entertainment, politics, sport and tech. The computer used for the analysis is a standard laptop with an Intel i-9 CPU at 5.1GHz, 64GB RAM, without the use of any additional hardware such as GPUs.

In Fig. 9, we show the similarities means per category across all pattern length for the three pairwise similarity comparisons, namely (a) ORIGINAL-SUSPICIOUS (blue), (b) ORIGINAL-REFERENCE (orange) and (c) SUSPICIOUS-REFERENCE (green). By observation, it is easy to note that ORIGINAL-SUSPICIOUS similarities are low on average compared to the other two, which is consistent with our previous work (K. F. Xylogiannopoulos et al., 2024a) for a completely different dataset. Another important observation is that the similarity ratios of SUSPICIOUS-REFERENCE are significantly higher than ORIGINAL-REFERENCE. Additionally, the one standard deviation spread is marginally overlapping in few cases. Therefore, the results seem to group separately between the two classes of similarities (for the majority of the articles).

This can further be observed in the frequency histogram of Fig. 10 and the boxplot diagrams of Fig. 11, for all categories and pattern length from three up to seven (longer pattern lengths are significantly skewed and cannot be visualized easily). In Fig. 9, it can be observed that descriptive statistics justify the expected results.

To evaluate the statistical significance of similarity differences between the classes, we performed a two-sample t-Test between the means of each combination of the three classes of similarities. For this, we calculated the means by applying random sampling with a 100 rounds bootstrapping, using the Central Limit Theorem (CLT). More precisely, we calculated the mean for each round by applying random sampling of 30 texts, using python standard random function with seed the value of each round (from zero to 99), per category. Then, the mean of the means is calculated, and because of the CLT, we know that it follows the Normal distribution.

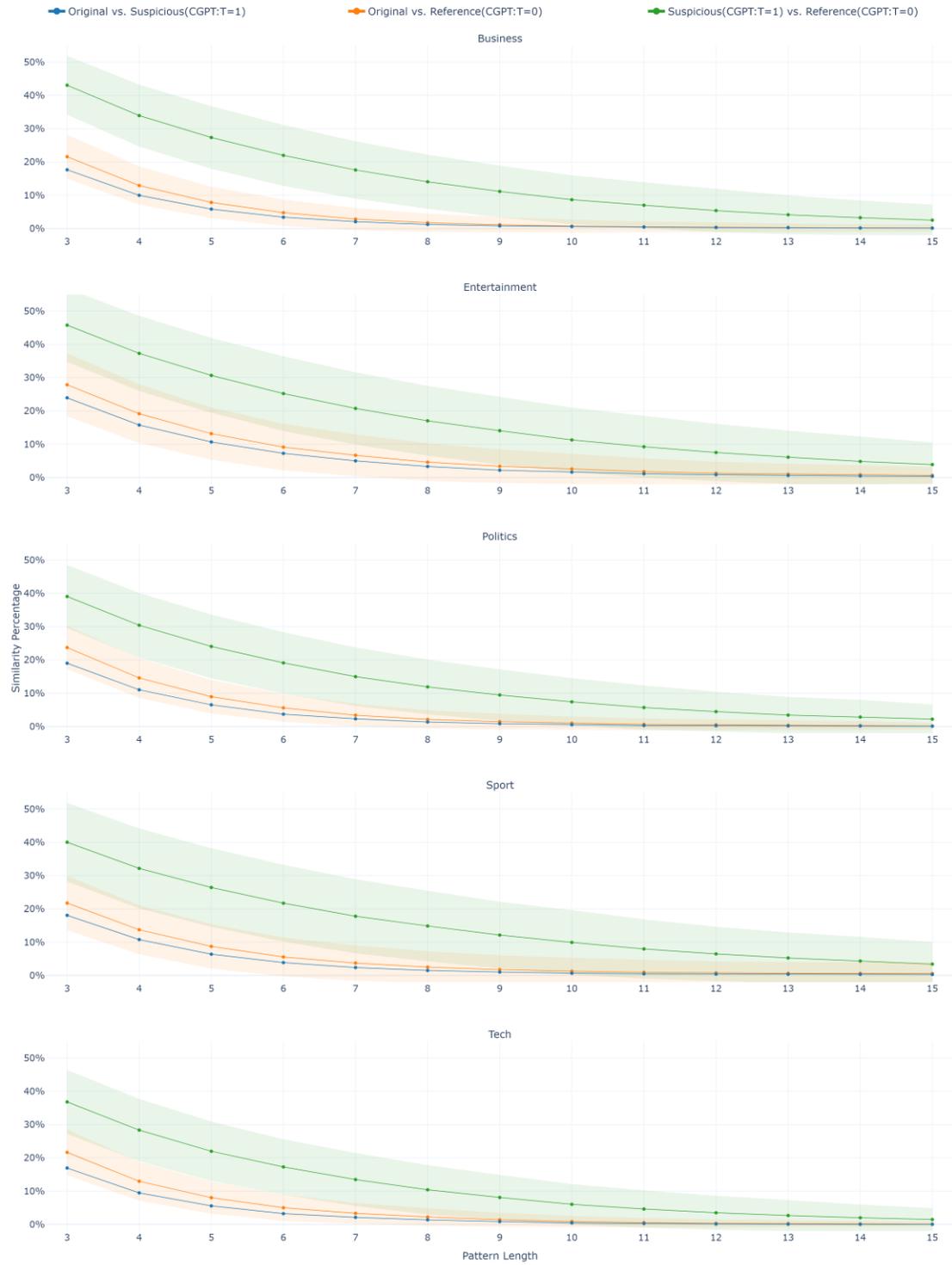

Fig. 9 Pairwise similarity comparisons namely (a) ORIGINAL-SUSPICIOUS (blue), (b) ORIGINAL-REFERENCE (orange) and (c) SUSPICIOUS-REFERENCE (green) for each of the five news category. Shaded areas represent one standard deviation of the data.

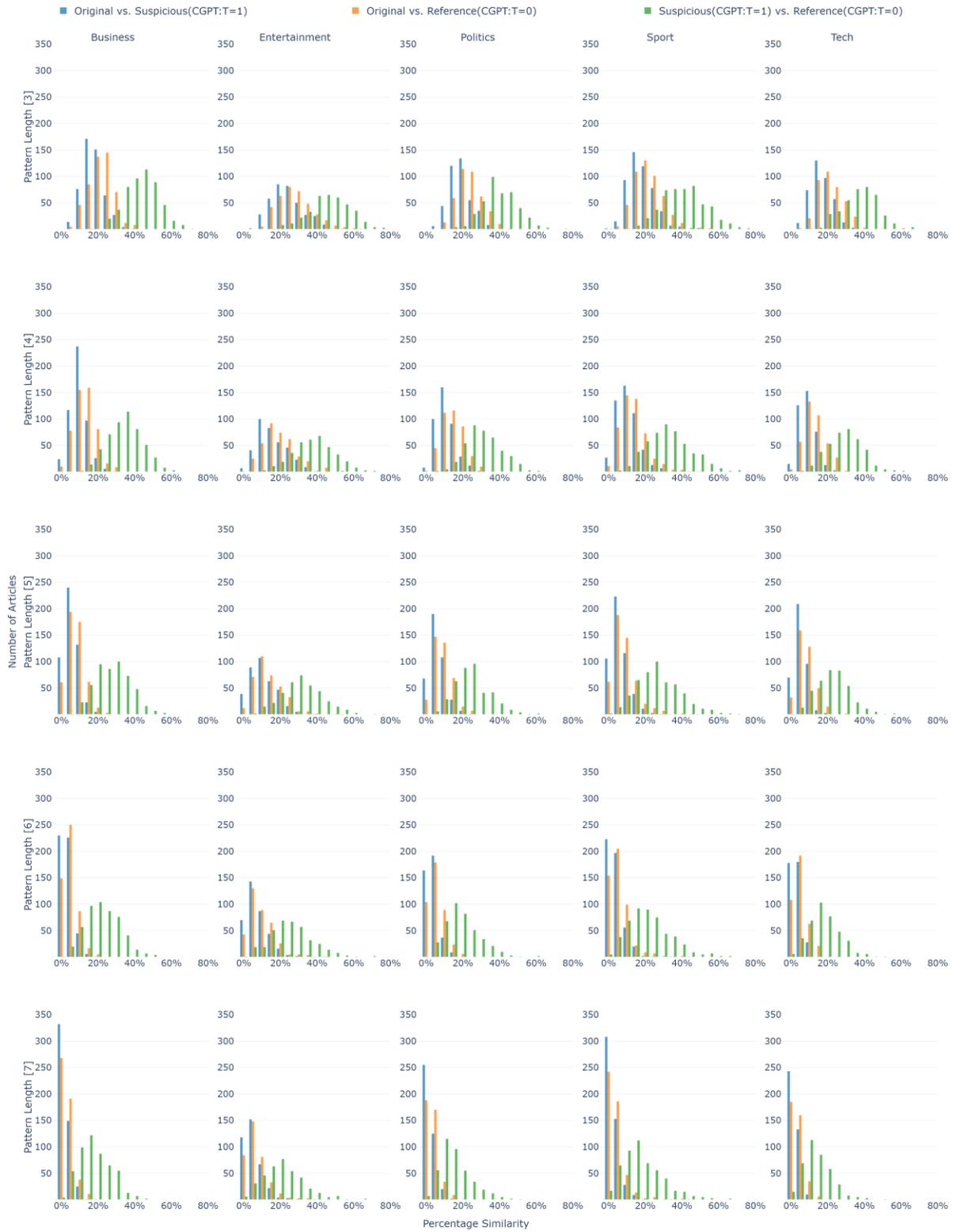

Fig. 10 Similarity histograms for each pair (a) ORIGINAL-SUSPICIOUS (blue), (b) ORIGINAL-REFERENCE (orange) and (c) SUSPICIOUS-REFERENCE (green) for each of the five news category.

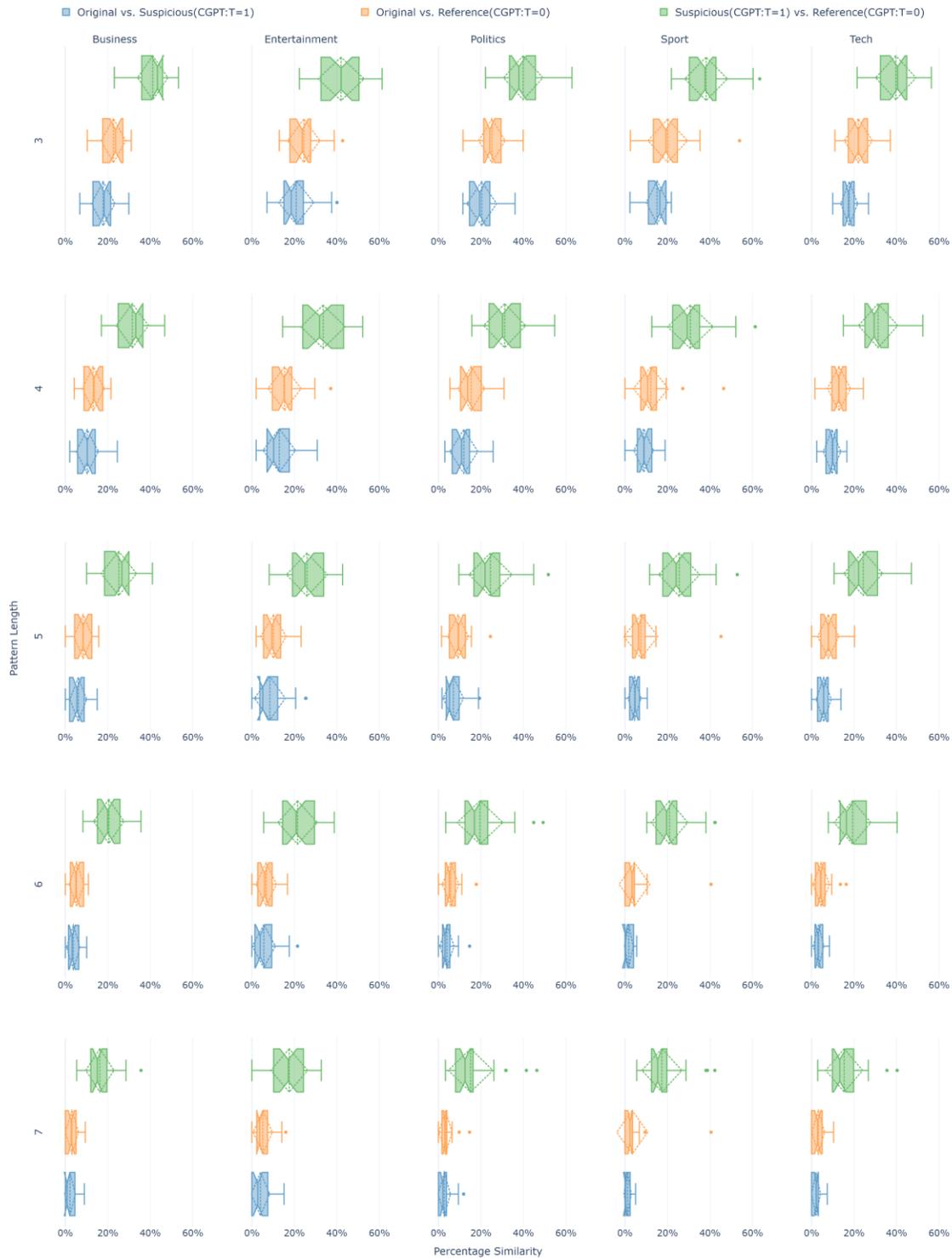

Fig. 11 Boxplot similarity diagrams for histograms for each pair (a) ORIGINAL-SUSPICIOUS (blue), (b) ORIGINAL-REFERENCE (orange) and (c) SUSPICIOUS-REFERENCE (green) for each of the five news category.

The three combinations checked are the ORIGINAL-SUSPICIOUS against the SUSPICIOUS-REFERENCE, the ORIGINAL-SUSPICIOUS against the ORIGINAL-REFERENCE and, finally, the ORIGINAL-REFERENCE against SUSPICIOUS-REFERENCE. In Table 2, we have the results of each test per category and pattern length. The t-statistic critical values of the test are presented, including the p-values for each test. Based on these values, we can infer that the similarity difference is statistically significant for SUSPICIOUS-REFERENCE against both ORIGINAL-SUSPICIOUS and ORIGINAL-REFERENCE. On the other side, for most of pattern length cases, there is no statistical significance between the similarities of ORIGINAL-SUSPICIOUS against ORIGINAL-REFERENCE for a significance level alpha 0.05. Even for the specific cases where the null hypothesis can be rejected, the p-values are significant higher compared to the rest of the comparison pairs. Therefore, we can infer that similarity comparisons are statistically significant when the common part in both pairs is one of the two paraphrases, i.e., Suspicious or Reference, in contrast to similarity comparisons where the common part is the Original.

Another very important observation is that the p-value, in most categories of the ORIGINAL-SUSPICIOUS against ORIGINAL-REFERENCE, starts very low, below significance level, and increases, above significance level, as the pattern length increases. Therefore, if we investigate only very short patterns it is very easy to fall into a pitfall and reject null hypothesis for this comparison. This observation demonstrates how important is to investigate not only short but also long pattern lengths, something that ARPaD and LERP-RSA can perform exceptionally easy. Moreover, the specific observation also reveals how the use of weights could significantly improve the classification whenever needed.

The above discussion serves as an indication that this methodology has the ability to distinguish between the Reference versus the Original and the Suspicious text. Practically, this result means that, on average, the similarity ratios do not overlap. It is also important to observe that ChatGPT paraphrases are significantly higher than the Original, with respect to the pattern length. These two observations serve as validators for our methodology as presented in Fig. 6 and Fig. 8. If the overall similarities sum for different pattern length is positive then a Suspicious text is classified as ChatGPT, otherwise it is classified as other, meaning that it is not ChatGPT but it can be human or any other source based, such as another LLM.

Table 2 Statistical comparison for each category and various pattern lengths

| Two-Sample t-Test between Percentage Similarities | | | | | | | |
|---|---|---|---|---|---|---|---|
| Dataset | Pattern Length | Original vs. Suspicious against Suspicious vs. Reference | | Original vs. Suspicious against Original vs. Reference | | Original vs. Reference against Suspicious vs. Reference | |
| | | t-statistic | p-value | t-statistic | p-value | t-statistic | p-value |
| Business | 3 | -13.939 | <10$^{-4}$ | -3.251 | 0.0019 | -10.839 | <10$^{-4}$ |
| | 4 | -12.393 | <10$^{-4}$ | -2.074 | 0.0425 | -10.767 | <10$^{-4}$ |
| | 5 | -11.135 | <10$^{-4}$ | -2.097 | 0.0404 | -9.598 | <10$^{-4}$ |
| | 6 | -11.562 | <10$^{-4}$ | -1.786 | 0.0794 | -10.344 | <10$^{-4}$ |
| | 7 | -10.672 | <10$^{-4}$ | -1.271 | 0.2087 | -9.676 | <10$^{-4}$ |
| Entertainment | 3 | -8.359 | <10$^{-4}$ | -1.740 | 0.0872 | -7.135 | <10$^{-4}$ |
| | 4 | -8.648 | <10$^{-4}$ | -1.243 | 0.2187 | -7.531 | <10$^{-4}$ |
| | 5 | -7.830 | <10$^{-4}$ | -0.978 | 0.3322 | -7.499 | <10$^{-4}$ |
| | 6 | -7.934 | <10$^{-4}$ | -0.825 | 0.4127 | -7.594 | <10$^{-4}$ |
| | 7 | -7.535 | <10$^{-4}$ | -0.934 | 0.3544 | -6.878 | <10$^{-4}$ |
| Politics | 3 | -9.069 | <10$^{-4}$ | -2.759 | 0.0077 | -7.130 | <10$^{-4}$ |
| | 4 | -8.739 | <10$^{-4}$ | -2.062 | 0.0437 | -7.270 | <10$^{-4}$ |
| | 5 | -8.413 | <10$^{-4}$ | -1.768 | 0.0823 | -7.327 | <10$^{-4}$ |
| | 6 | -7.432 | <10$^{-4}$ | -1.831 | 0.0722 | -6.513 | <10$^{-4}$ |
| | 7 | -6.111 | <10$^{-4}$ | -0.719 | 0.4750 | -5.798 | <10$^{-4}$ |
| Sport | 3 | -11.096 | <10$^{-4}$ | -2.555 | 0.0133 | -7.152 | <10$^{-4}$ |
| | 4 | -10.133 | <10$^{-4}$ | -1.920 | 0.0598 | -7.410 | <10$^{-4}$ |
| | 5 | -11.213 | <10$^{-4}$ | -1.909 | 0.0612 | -7.753 | <10$^{-4}$ |
| | 6 | -11.702 | <10$^{-4}$ | -1.923 | 0.0594 | -7.933 | <10$^{-4}$ |
| | 7 | -9.120 | <10$^{-4}$ | -1.557 | 0.1248 | -6.348 | <10$^{-4}$ |
| Tech | 3 | -11.459 | <10$^{-4}$ | -3.012 | 0.0038 | -8.199 | <10$^{-4}$ |
| | 4 | -11.560 | <10$^{-4}$ | -2.596 | 0.0119 | -9.246 | <10$^{-4}$ |
| | 5 | -10.206 | <10$^{-4}$ | -1.885 | 0.0645 | -8.513 | <10$^{-4}$ |
| | 6 | -9.586 | <10$^{-4}$ | -1.444 | 0.1541 | -8.377 | <10$^{-4}$ |
| | 7 | -7.981 | <10$^{-4}$ | -1.581 | 0.1193 | -7.152 | <10$^{-4}$ |

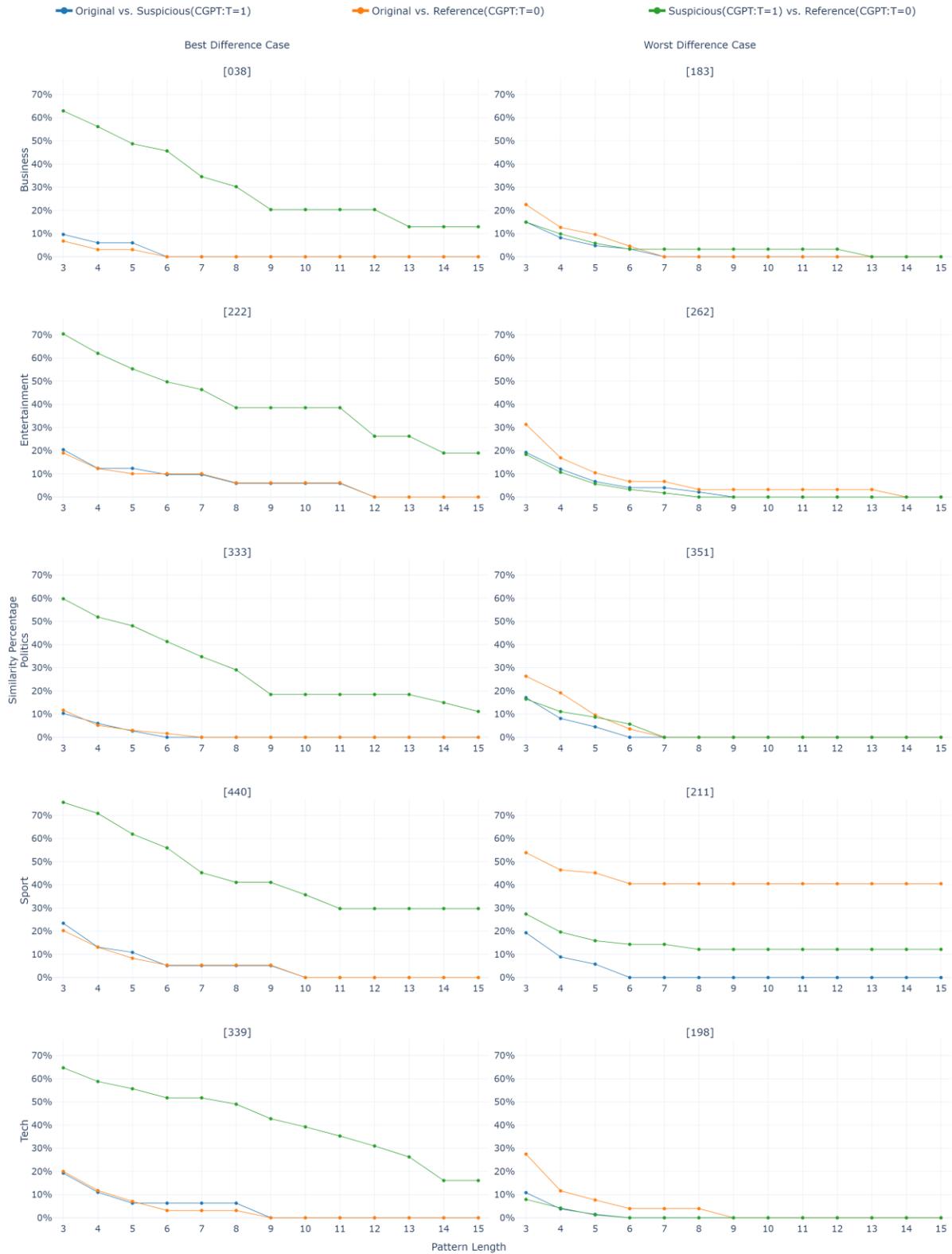

Fig. 12 Examples of best (highest possible similarity difference) and worse (lowest possible similarity difference) of news articles for pattern length three. Bracketed numbers refer to the article ID from the original dataset.

In Fig. 12, we provide the best and worst difference similarity ratio examples for each category for pattern length equal to three words. The difference is always calculated between SUSPICIOUS-REFERENCE (green) and ORIGINAL-REFERENCE (orange). We can observe how the text similarities for each case can vary from very wide (first column) to very narrow or even negative (second column) for patterns length three. The importance of performing such similarities checks for different pattern length, as previously discussed, can be shown in the worst case of the Business category. Although the similarities ratios between ORIGINAL-REFERENCE are higher for the first four pattern lengths (3-6), no common patterns exist for lengths seven and above. However, longer common patterns (7-12) exist between SUSPICIOUS-REFERENCE. Therefore, since the score for the shorter pattern lengths is minus four and for the rest is plus six, the overall score is plus 2 and the Suspicious text is classified as ChatGPT.

In Fig. 13 through Fig. 17, we have diagrams that each word is presented as a dash in the plot while each colored block represents the corresponding similar word. Each figure is partitioned (based on the colors) in three groups with regard the texts compared (y-axes labels). For example, the first two lines present the comparison between the Original and the Suspicious texts, which also have different lengths as we can observe from the dashes. There are also four blue colored blocks presenting similar words between them. Of course, patterns can overlap forming longer blocks of similarities in the plot and, additionally, it can also be observed that these blocks do not occur at the same, or close enough, positions in each text. This is very important because ARPaD can detect patterns regardless position in the different texts. For the presented examples in the figures, the full texts with marked similarities are also presented in Appendix I, for direct comparison purposes.

It is also noteworthy that the best cases occur between texts of approximately the same length while the worst cases occur between texts with significantly different lengths. This is related to the fact that ChatGPT API cannot create paraphrases for longer texts (>500 words) and can be easily observed by the absence or limited number of common patterns of the Original text above the threshold of 500 words. For example, in Fig. 13, where the article 183 of the business category is examined as the worst category case for pattern length 3, word counts for each of the Original, Suspicious and Reference are 813, 354 and 395 respectively. The corresponding similarities between REFERENCE-ORIGINAL and REFERENCE-SUSPICIOUS are 22.53% and 14.94%, since for the first couple there are 89 similar words for pattern length 3 and for the second couple there are 59 similar words. However, there is a large block of common text with length 12 words, which explains what mentioned above for the similar patterns of length 3-12 (overlapping). Such kind of long patterns can be defined as "signatures" of the text, which can reveal possible ChatGPT paraphrases and can be captured only by detecting and analyzing variable pattern lengths.

The occurrence of such worst cases is expected, as described in previous section, since the significantly diverse text lengths can affect the similarity ratios calculated based on the text's length. However, even in such cases the proposed methodology can detect suspicious text and classify them correctly despite this peculiarity. Furthermore, it will be presented in the next section, with the performance metrics of the methodology, that the results are very good regardless the fact that many texts have lengths significantly higher than the ChatGPT limitation, as we have observed in Section 4.1.1 (Fig. 1). Therefore, there was no reason to add complexity using normalization, something that, of course, can be used in other cases where it might be useful to improve the performance without overfitting.

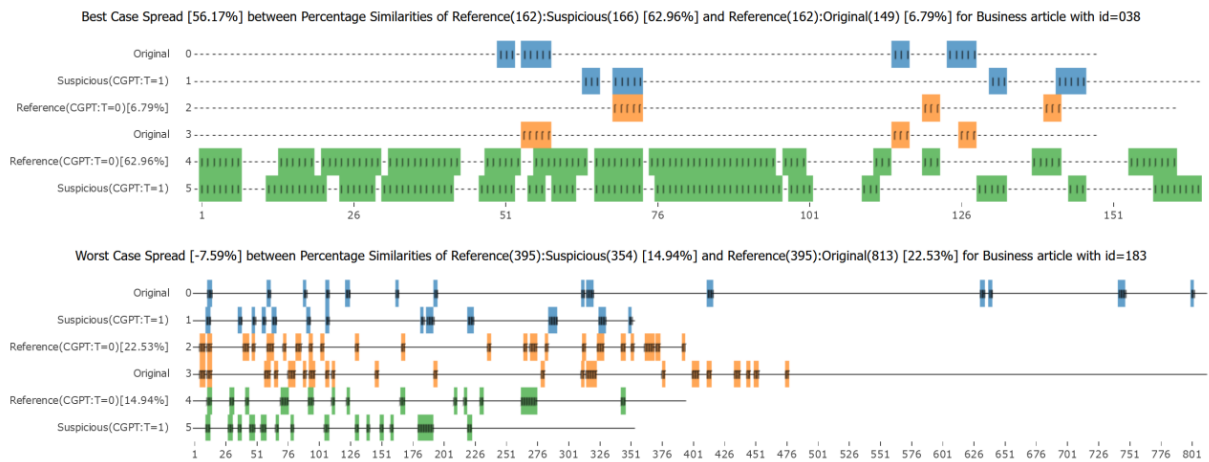

*Fig. 13 Examples of best and worst case ratio spread similarities for pattern (colored blocks) length three by word (dash) representation for Business category.*

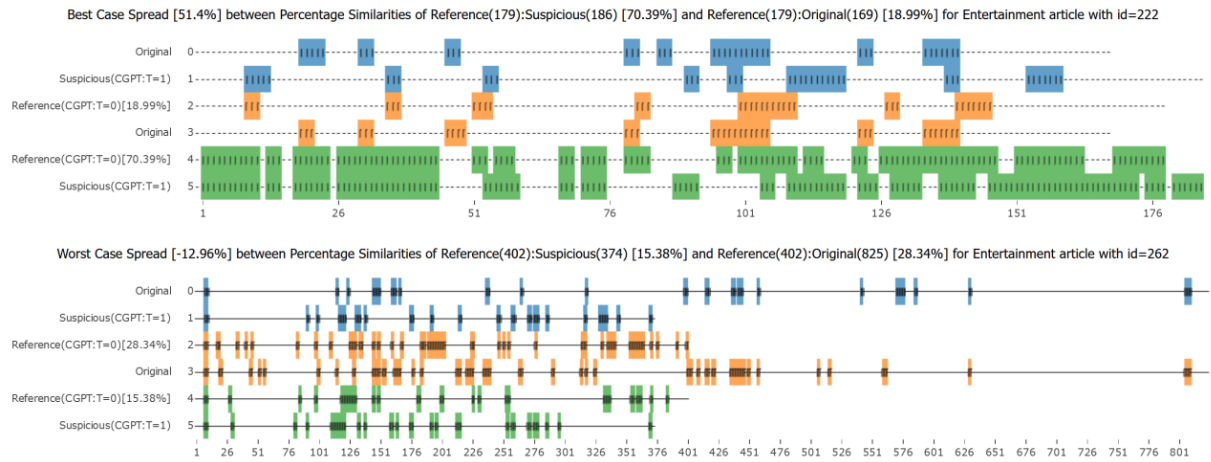

Fig. 14 Examples of best and worst case ratio spread similarities for pattern (colored blocks) length three by word (dash) representation for Business category.

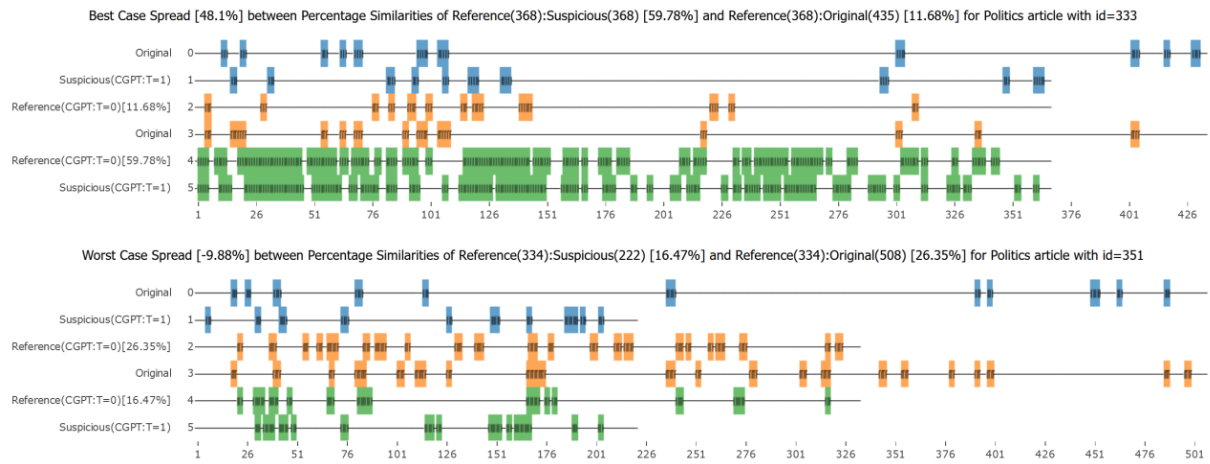

Fig. 15 Examples of best and worst case ratio spread similarities for pattern (colored blocks) length three by word (dash) representation for Politics category.

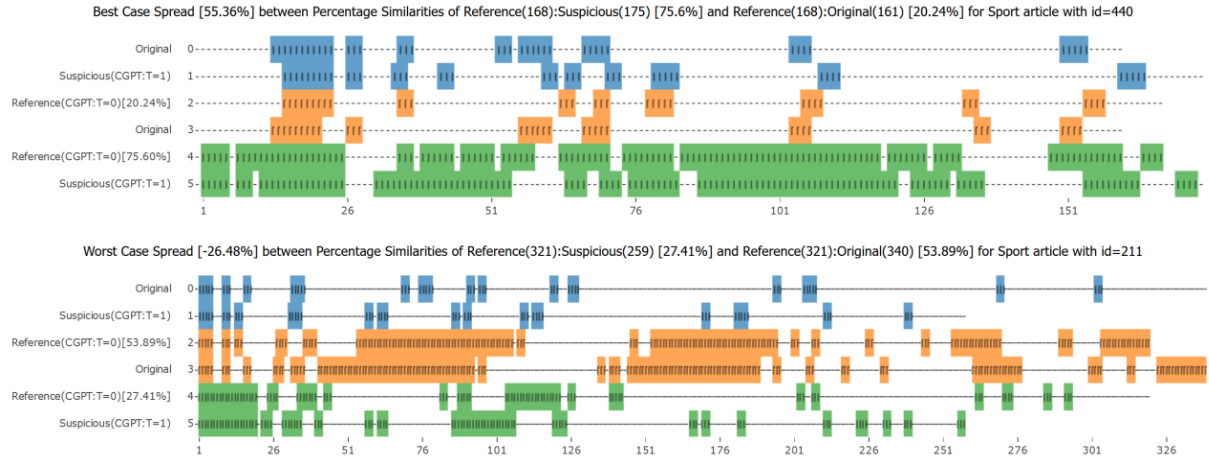

*Fig. 16 Examples of best and worst case ratio spread similarities for pattern (colored blocks) length three by word (dash) representation for Sport category.*

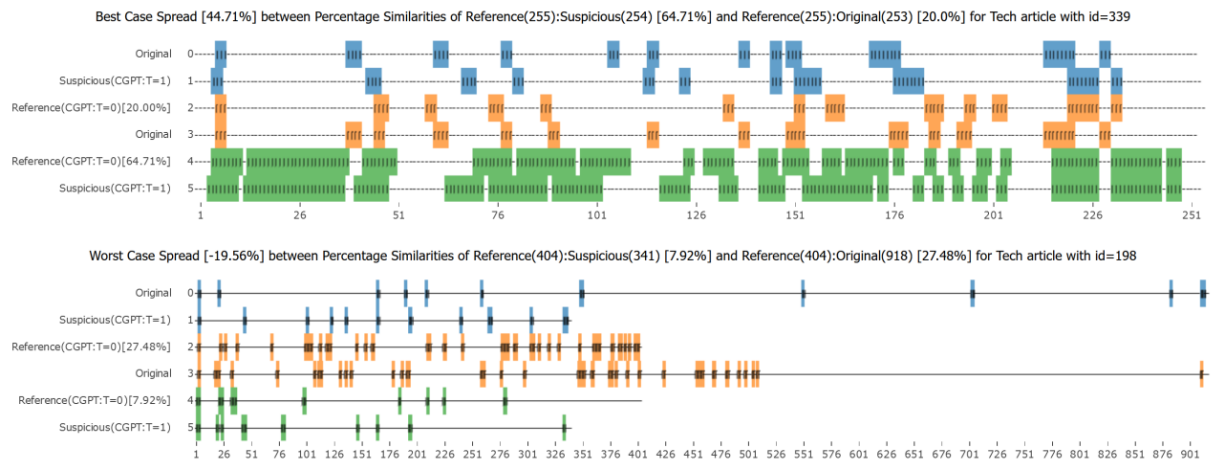

*Fig. 17 Examples of best and worst case ratio spread similarities for pattern (colored blocks) length three by word (dash) representation for Tech category.*

### 4.2. Dataset Expansion for Methodology Evaluation

Before we proceed with an expanded version of our dataset to have a complete evaluation of the methodology, it is important to define a new type of input text as follows:

**Definition 4**: We define as "Other" a text of the same subject and/or topic as the Original text, which has been produced by any other way than ChatGPT, human or artificial.

In this paper, the Other text is created with the use of another LLM (Gemini) to introduce an additional paraphrase, which is not ChatGPT based, and it will be used as the non-ChatGPT version of the text. Although the Other text in a real-world scenario could be original articles of the same topic from other news organizations and, therefore, human created, unfortunately, the dataset is very large to create manually, human based, paraphrases to test it and, furthermore, there is no data from another news agency for the exact same news that could be used. However, this extra paraphrase (by Gemini) is very important to balance our dataset with negative inputs and be able to produce True Negative and False Positive results. A dataset including only ChatGPT texts would be completely imbalanced and it would not be possible to evaluate the methodology with regard to the research objectives (Section 1.2). Therefore, it is not possible to evaluate how the methodology will behave when a suspicious paraphrased text is a non-ChatGPT created paraphrase unless such text is introduced, even with the help of another LLM, since the scope of the research is to produce a methodology to explicitly detect ChatGPT paraphrases and not simply distinguish between human and AI.

Additionally, it is important to mention that due to technological limitations of Gemini to produce meaningful paraphrases for the full dataset, the new dataset of original texts, ChatGPT and Gemini paraphrases has been transformed to 508 business, 368 entertainment, 402 politics, 501 sport and 386 tech, in total 2,165 instead of 2,224 of the original dataset.

So far, we have not used the extra text created with Gemini. As we can see in Fig. 18, the overall average similarities between the REFERENCE-OTHER are lower than the REFERENCE-ORIGINAL. This also validates our initial assumption that the two LLMs produce significantly different text for reasons, such as, different training datasets, input parameters, size, etc. Furthermore, it allows to enrich the initial dataset of the BBC news with additional, similar, news text, which although they are AI created (Gemini), yet, for simulation purposes works very well when we want to detect explicitly ChatGPT paraphrases.

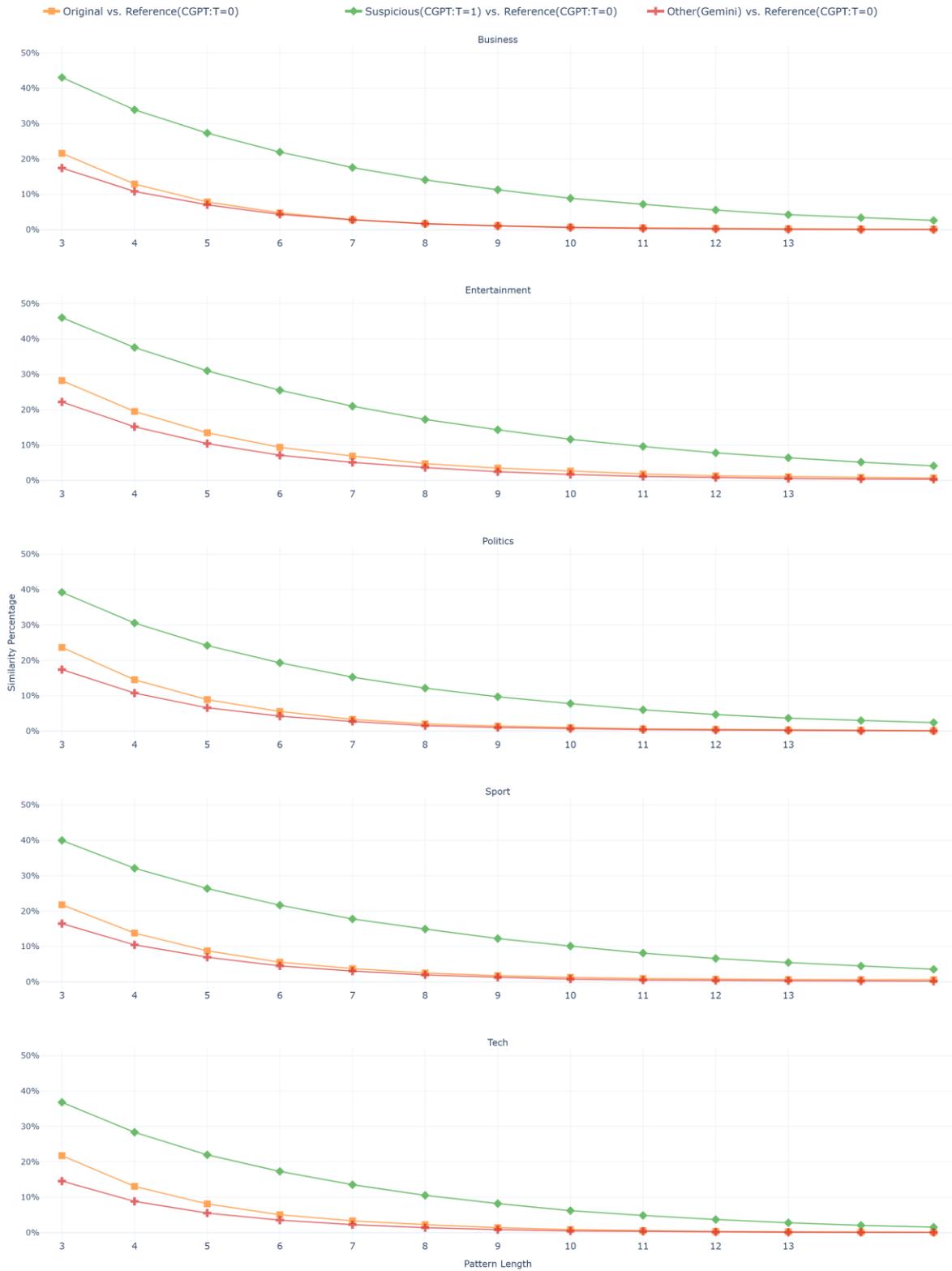

Fig. 18 Pairwise similarity comparisons namely (a) ORIGINAL-REFERENCE (orange), (b) SUSPICIOUS-REFERENCE (green) and (c) OTHER-REFERENCE (red) for each of the five news category.

Using the same example of Section 3.2 (Fig. 4), we can observe how different the Other text (second column) is and how it has much fewer common patterns with the Original text (first column) and Reference text (third column) and can further explain the results presented in Fig. 19. The first column depicting the similarities between the Reference and the Original on the Original text is the same as in Fig. 4. However, in the second column where the similarities between the Reference and the Other are presented on the Other, we can observe that common patterns are rare and limited in names and places. This is further observable in the third column where all common patterns are presented on the Reference text and the patterns between the Original and Other, although similar in size, are completely different, except some trivial patterns.

*Fig. 19 An example of an Original text (left) an Other text (center) and a Reference text(right) along with their pariwise similarity patterns shown in different colors.*

## 5. Discussion

### 5.1. Performance and Comparison with other tools

Based on the analysis performed in the previous section, we can observe the results in the Confusion Matrices per news category of Fig. 20.

|   | Business | | Entertainment | | Politics | | Sport | | Tech | |
|---|---|---|---|---|---|---|---|---|---|---|
| | Predicted | | Predicted | | Predicted | | Predicted | | Predicted | |
| | ChatGPT-4 | Other | ChatGPT-4 | Other | ChatGPT-4 | Other | ChatGPT-4 | Other | ChatGPT-4 | Other |
| ChatGPT-4 | 503 | 5 | 353 | 15 | 390 | 12 | 470 | 31 | 367 | 19 |
| Other | 20 | 488 | 8 | 360 | 14 | 388 | 23 | 478 | 16 | 370 |

*Fig. 20 Confusions matrices per category for Phase II classification.*

These results are summarized in Table 3 using standard metrics, i.e., accuracy, precision, sensitivity, specificity and the overall F1-score. The overall performance for the full dataset can be depicted in the last column.

*Table 3 Overall Results Performance by Confusion Matrices Metrics*

| Dataset → <br> Metric (%) ↓ | Business | Entertainment | Politics | Sport | Tech | Overall |
|---|---|---|---|---|---|---|
| Accuracy | 97.54 | 96.88 | 96.77 | 94.61 | 95.47 | 96.23 |
| Precision | 96.18 | 97.78 | 96.53 | 95.33 | 95.82 | 96.25 |
| Sensitivity | 99.02 | 95.92 | 97.01 | 93.81 | 95.08 | 96.21 |
| Specificity | 96.06 | 97.83 | 96.52 | 95.41 | 95.85 | 96.25 |
| F1 | 97.58 | 96.84 | 96.77 | 94.57 | 95.45 | 96.23 |

Next, we perform a comparison with existing tools, and for this reason we have used RADAR, an open source AI detection algorithm (Hu et al., 2023). Unfortunately, as most of the AI detection software, to the best of our knowledge, RADAR can classify texts only as AI and Human and cannot distinguish between LLMs. Another important reason of using the specific tool is that is the only one, to the best of our knowledge, to provide an intuitive API that can be used to automatically process the thousands of texts. Using any other tool it would make the reproduction and validation of the presented research work practically impossible, since most the available tools are either not free or they do not provide an API for such purposes, or both.

Since RADAR cannot distinguish and detect a specific LLM, therefore, a direct comparison with our methodology is not feasible due to the nature of the dataset. However, what it can be compared is the ability of RADAR to classify the ChatGPT as AI or human. This kind of comparison is adequate because if it fails to classify ChatGPT as AI then definitely it is not possible to distinguish between the two LLMs. In such case, our decision to use Gemini for the other text, instead of human, for our methodology valication is fully justified.

The results of such analysis can be observed in the next figure (Fig. 21) showing that practically RADAR fails to detect ChatGPT paraphrased news articles and classifies almost all of them as Human.

| Business | | | Entertainment | | | Politics | | | Sport | | | Tech | | |
|---|---|---|---|---|---|---|---|---|---|---|---|---|---|---|
| | Predicted | | | Predicted | | | Predicted | | | Predicted | | | Predicted | |
| | AI (ChatGPT-4) | Human | | AI (ChatGPT-4) | Human | | AI (ChatGPT-4) | Human | | AI (ChatGPT-4) | Human | | AI (ChatGPT-4) | Human |
| Actual AI (ChatGPT-4) | 2 | 507 | Actual AI (ChatGPT-4) | 6 | 380 | Actual AI (ChatGPT-4) | 10 | 406 | Actual AI (ChatGPT-4) | 53 | 458 | Actual AI (ChatGPT-4) | 7 | 394 |

*Fig. 21 RADAR performance per category for True Positive/False Negative classifications*

Overall, the proposed methodology is robust since it is performed in an one-to-one basis, with no need of complicated machine learning or deep learning models. Furthermore, because of ARPaD and LERP-RSA, the worst case computational time and space complexity is $O(mnlogn)$ where $m$ is the number of texts to analyze (two for the first phase and three for the second phase; practically a constant) and $n$ is the maximum length between the compared texts. From a computational perspective, the overall worst-case time and space complexity is practically constant, since the number of articles $m$ is fixed (three) and the length of articles is also fixed, in the sense that it cannot grow beyond a limited number of few hundred words. It is obvious that the theoretical and practical computational cost, is significantly low because for our approach and it can outperform other tools due to its simplicity and straightforward implementation. Any attempt to use machine or even worse deep learning approaches could lead to significant costs.

Instead of using a black-box machine or deep learning approach, a direct classification is performed based on pattern similarities between Original and ChatGPT Suspicious paraphrase texts with the use of an alternative ChatGPT paraphrase. A direct effect of this methodology is its high efficacy and very low cost for implementation since it can be performed in per use case basis instead of a generic model that would also need additional cost for constant update and maintenance.

### 5.2. Practical and theoretical implications

In addition to what presented in the previous section, the proposed methodology makes possible the direct identification of the underlying LLM technology that creates the paraphrases and classify them as ChatGPT instead of a generic AI vs Human classification scheme. This could be of extremely importance, e.g., for legal cases where consistent abuse of copyright infringement of news agencies organization may occur from LLMs.

This diversification between ChatGPT paraphrases and original articles has been tested through the simulation of the ChatGPT and Gemini texts and the very high and well-balanced metrics scores achieved. Although this is the purpose of the methodology, i.e., detect ChatGPT paraphrases, yet, this could be classified as a possible disadvantage since other LLMs can be directly classified as non-human, but this is another use case and in this case a new dataset is needed, as explicitly described in Section 3.1, with very specific attributes and furthermore with additional sources of human texts. Furthermore, simply classifying paraphrases as AI may have no usefulness in the real (business) world since the actual infringer cannot be detected. We hope that this work will stimulate more research around algorithmic text similarity.

The methodology has been described in detail with additional features, which although not used for the specific dataset due to its already high performance, can help improve results. Such examples are the text preprocessing of apostrophes, which could help to acquire higher scores, normalization of text similarities with regards to the length of the compared texts, the use of different weights to assign different importance to longer and, thus, rarer to occur naturally patterns, etc.

## 6. Conclusion

In this study, we present a methodology that addresses the business case of news agency organizations copyright infringement with the use of AI. More specifically, we demonstrate the effectiveness of a text-based pattern similarity methodology in detecting ChatGPT-paraphrased news. We propose an algorithmic approach capable of not only identifying whether an article has been paraphrased by AI tools but more importantly tracing the original ChatGPT LLM source. Our method is evaluated using a benchmark dataset specifically designed for this task, which includes real BBC articles combined with ChatGPT-generated paraphrases and Gemini, for simulation purposes only, to identify potential ChatGPT paraphrases. The results show that our approach achieves over 95% accuracy, precision, sensitivity, and specificity in detecting ChatGPT-assisted paraphrased articles. Unlike traditional machine or deep learning models trained on specific datasets, our method relies on direct pattern similarity comparison which make it robust and cost-effective.

In future work, we will attempt to find or generate a more applicable dataset which will include articles from multiple news agency organizations for the same topic. Additionally, as more LLMs appear that provide APIs, we intend to generalize our methodology to detect and attribute the suspicious text to the corresponding LLM which produced it. In that effort, we expect to formalize and extend the use of weights to distinguish between different LLMs and improve results that may demonstrate performance lower than 90%.

# APPENDIX I

## BBC Article (Business) 038

| Original - Reference | Suspicious - Reference | Reference - Original & Suspicious | Other - Reference |
|---|---|---|---|
| German telecoms firm Deutsche Telekom saw strong fourth quarter profits on the back of upbeat US mobile earnings and better-than-expected asset sales. Net profit came in at 1.4bn euros (£960m; $1.85bn), a dramatic change from the loss of 364m euros in 2003. Sales rose 2.8% to 14.96bn euros. Sales of stakes in firms including Russia's OAO Mobile Telesystems raised 1.17bn euros. This was more than expected and helped to bring debt down to 35.8bn euros. A year ago, debt was more than 11bn euros higher. T-Mobile USA, the company's American mobile business, made a strong contribution to profits. "It's a seminal achievement that they cut debt so low. That gives them some head room to invest in growth now," said Hannes Wittig, telecoms analyst at Dresdner Kleinwort Wasserstein. The company also said it would resume paying a dividend, after two years in which it focused on cutting debt. | Deutsche Telekom, a German telecommunications company, experienced robust profit growth in the fourth quarter, driven by successful operations in the US mobile market and the sale of assets that exceeded expectations. The company reported a net profit of 1.4 billion euros, a significant turnaround from a 364 million euro loss in 2003. Revenues increased by 2.8% to reach 14.96 billion euros. The disposal of stakes in various companies, including Russia's OAO Mobile Telesystems, brought in 1.17 billion euros, contributing to a reduction in the firm's debt to 35.8 billion euros, which is over 11 billion euros lower than the previous year. The performance of T-Mobile USA was notably influential in the profit rise. "They've achieved a milestone in reducing their debt to such an extent. Now they have the flexibility to invest in expanding their business," stated Hannes Wittig, a telecommunications analyst at Dresdner Kleinwort Wasserstein. Additionally, Deutsche Telekom announced its plans to start paying dividends again after a two-year period of prioritizing debt reduction. | Deutsche Telekom, a German telecommunications company, experienced a significant increase in profits during the fourth quarter, driven by successful mobile operations in the US and the sale of assets that surpassed expectations. The company reported a net profit of 1.4 billion euros, a stark turnaround from the 364 million euro loss in 2003. Revenue climbed by 2.8% to 14.96 billion euros. The sale of shares in various companies, including Russia's OAO Mobile Telesystems, generated 1.17 billion euros, contributing to a reduction in the firm's debt to 35.8 billion euros, which is over 11 billion euros less than the previous year. T-Mobile USA, Deutsche Telekom's American mobile division, played a significant role in the profit surge. Analyst Hannes Wittig from Dresdner Kleinwort Wasserstein remarked on the importance of the company's debt reduction, noting it provides Deutsche Telekom with the flexibility to invest in future growth. Additionally, the company announced plans to reinstate dividend payments after a two-year period of prioritizing debt reduction. | Deutsche Telekom's fourth quarter profits surged thanks to robust mobile earnings in the United States and higher-than-anticipated asset sales. Amidst these gains, the telecoms giant witnessed a significant turnaround, transforming a 2003 loss of 364 million euros into a net profit of 1.4 billion euros. Sales experienced a 2.8% uptick, reaching 14.96 billion euros, showcasing the company's overall financial recovery. Notably, asset sales, including stakes in OAO Mobile Telesystems in Russia, yielded 1.17 billion euros, far surpassing initial expectations. This financial windfall allowed Deutsche Telekom to reduce its debt burden by 11 billion euros, bringing it down to 35.8 billion euros. Furthermore, T-Mobile USA, the company's American mobile business, played a crucial role in driving profits. To mark this financial success, Deutsche Telekom announced it would resume paying dividends, signaling its intention to prioritize growth after two years of focused debt reduction. Hannes Wittig, a telecoms analyst at Dresdner Kleinwort Wasserstein, hailed the company's achievement in slashing debt, providing significant room for future investment. |

# BBC Article (Business) 183

| Original - Reference | Suspicious - Reference | Reference - Original & Suspicious | Other - Reference |
|---|---|---|---|
| As European leaders gather in Rome on Friday to sign the new EU constitution, many companies will be focusing on matters much closer to home namely how to stay in business. Lille is a popular tourist destination for Britons who want a taste of France at the weekend. But how many tourists look at the impressively grand Victorian Chambre de Commerce, which stands beside the Opera House, and consider that it was built like the town halls in many northern English towns on the wealth created by coal, steel and textiles? Like northern England and industrial Scotland, those industries have been in long term decline the last coal pit closed in 1990. Beck-Crespel is a specialist steel firm in Armentieres, about 20 miles from Lille. The company has not laid off a worker since 1945. It specialises in making bolts and fixings for power stations and the oil industry, but not many of those are being built in Europe these days. Director Hugues Charbonnier says he is under pressure because factories in the Far East are able to make some of his output more cheaply, while his key markets are now in China and India. \"In our business the market is absolutely global, you can not imagine living with our size (of business) even within an enlarged European Union, (if we did that) we would need not 350 people but perhaps just 150 or 200,\" he says. It isn't just globalisation that is hurting; the law in France means workers are paid for a 39 hour week even though they work just 35 hours. But at least there is still a steel industry. Coal has now totally vanished and textiles are struggling. New business has been attracted, but not enough to make up the difference. That is one reason why people here are not great fans of the EU, says Frederic Sawicki, a politics lecturer at the University of Lille. \"In the region today the unemployment rate is 12%, in some areas it is 15%. They don't see what Europe is doing for them, so there is a kind of euro scepticism, especially in the working classes,\" he says. Which is strange because Lille is at the crossroads of Europe if anywhere should be benefiting from the euro it is here. The euro was designed to increase trade within the eurozone, but the biggest increase in trade has been with the rest of the world. Much of that trade passes through the world's largest port, Rotterdam, in Holland, home to specialist crane maker Huisman Itrec. Its cranes help build oil rigs and lifted the sunken Russian submarine Kursk from the sea bed, but Huisman Itrec is now setting up a factory in China, where costs are cheaper and its main customers are closer. Boss Henk Addink blames the low growth rate in Europe for the lack of orders closer to home. \"In the US growth is something like 6%, in China they are estimating 15%, and in the EU it is more or less 1%,\" he says. Mr Addink blames the euro for stifling demand. He much preferred the old currencies of Europe, which moved in relation to each country's economic performance. In Germany, industry is exporting more these days, but the economy as a whole is once again mired in slow growth and high unemployment. Growth is likely to peak this year at just under 2%. In Britain that would be a bad year; in Germany it is one of the best in recent years. With Germany making up a third of the eurozone's economy, this is a major problem. If Germany doesn't once again become the powerhouse of Europe, growth across the bloc is never going to be as strong as it could be. However, at one factory near the Dutch border things are changing. The Siemens plant at Boscholt makes cordless phones and employs 2,000 staff. Staff have started working an extra four hours a week for no extra pay, after Siemens threatened to take the factory and their jobs to Hungary. Factory manager Herbert Stueker says that he now hopes to increase productivity \"by nearly 30%\". But Germany needs much more reform if all its industry is to compete with places such Hungary or China. The Government is reforming the labour market and cutting the generous unemployment system, but the real solution is to cut the wages of low skilled workers, says Helmut Schneider, director of the Institute for the Study of Labour at Bonn University. \"Labour is too costly in Germany, especially for the low skilled labour and this is the main problem. If we could solve that problem we could cut unemployment by half,\" he says. The EU set itself the target of being the most efficient economy in the world by 2010. Four years into that process, and the target seems further away than ever. | As European government heads converge in Rome to ratify the new EU constitution on Friday, numerous businesses will be preoccupied with immediate survival concerns. Lille, a French city favored by British weekenders, boasts the impressive Chambre de Commerce, harking back to the region's industrial wealth from coal, steel, and textiles—industries now in decline, with the last coal mine shuttering in 1990. Around 20 miles from Lille in Armentieres, steel specialist Beck-Crespel, which hasn't downsized since 1945, primarily produces bolts and fixtures for energy sectors, finds fewer such projects in Europe. Director Hugues Charbonnier points to competition from Asia and a shift in key markets to China and India, noting the necessity for his business to operate globally. With the European market insufficient, Charbonnier suggests a smaller workforce might suffice. French labor laws also pose a challenge, mandating pay for a 39-hour week for only 35 hours worked. The regional steel industry persists despite the disappearance of coal and textile industries facing difficulties. EUR skeptic sentiment echoes in regions like Lille, amplified by high unemployment and little perceived EU benefit, according to Frederic Sawicki, a political science lecturer at the University of Lille. Lille's strategic European location contrasts its economic struggles, despite the euro's trade enhancement goals, with greater global trade focus. Huisman Itrec, a Dutch crane manufacturer situated in the world's largest port, Rotterdam, is relocating production to China due to Europe's tepid growth and proximity to key markets. Huisman Itrec's Henk Addink attributes scant local orders to Europe's low growth rates compared to the U.S. and China, and laments the rigidity of the euro has brought to European economies. Germany's industry faces similar issues—a meager growth with considerable unemployment. Even with some firms like the Siemens plant near the Dutch border, where workers agreed to extra hours without pay bumps to prevent relocation, the country still needs larger reforms, particularly wage reductions for low-skilled labor as suggested by Helmut Schneider from the Bonn University's Institute for the Study of Labour. Despite the EU's ambition to become the leading global economy by 2010, this goal seems increasingly elusive only four years into the challenge. | As European officials convene in Rome on Friday to formalize the new EU constitution, numerous businesses are preoccupied with more immediate concerns, such as their survival. Lille, a city favored by British tourists for short French getaways, boasts the grand Victorian Chambre de Commerce next to the Opera House. Few visitors may realize that this edifice, similar to town halls in many northern English cities, was funded by the wealth from coal, steel, and textile industries. These sectors, once thriving in northern England and industrial Scotland, have been declining for years, with the last coal mine in the area shutting down in 1990. Beck-Crespel, a specialized steel company located near Lille in Armentieres, has managed to avoid layoffs since 1945. The firm produces bolts and fasteners for power plants and the oil sector, yet the demand for such constructions in Europe has dwindled. Hugues Charbonnier, the company's director, faces competition from Asian factories that can produce some items at lower costs, and he now looks to China and India for his primary markets. Charbonnier points out the necessity of operating on a global scale, stating that to survive solely within the EU, his company would need to significantly reduce its workforce. The challenges aren't solely due to globalization; French labor laws mandate that employees are compensated for a 39-hour workweek even if they only work 35 hours. While the steel industry persists, coal has disappeared and textiles are faltering. Despite new businesses being drawn to the area, they haven't compensated for the losses. This economic situation fuels local skepticism towards the EU, according to Frederic Sawicki, a political science lecturer at the University of Lille. He notes the high unemployment rates in the region and the perception that the EU isn't delivering benefits, particularly among the working class. Lille's strategic location at the heart of Europe would suggest it should benefit from the euro, which was intended to boost intra-eurozone trade. However, trade with the rest of the world has seen the most significant increase. This global trade often passes through Rotterdam, the world's largest port, where the crane manufacturer Huisman Itrec is based. Despite its impressive projects, such as lifting the sunken Russian submarine Kursk, Huisman Itrec is establishing a factory in China to be closer to its customers and to save on costs. Henk Addink, the company's head, attributes the lack of | As European leaders prepare to sign the new EU constitution in Rome, businesses across Europe are preoccupied with staying afloat. In France, the once-thriving industries of coal, steel, and textiles have declined, leaving cities like Lille struggling economically. Businesses like Beck-Crespel, a steel firm, face growing pressure from cheaper imports from the Far East and a shift in demand to markets like China and India. Lille's unemployment rate remains high, causing skepticism about the benefits of the EU among working-class citizens. While trade within the Eurozone increased after the introduction of the euro, most trade growth has come from outside the bloc, specifically through ports like Rotterdam in Holland. Huisman Itrec, a crane maker based in Rotterdam, attributes its decline in European orders to the low growth rate. They have opened a factory in China, closer to their main customers and where operating costs are lower. Germany, representing a third of the Eurozone's economy, is experiencing economic challenges. Despite increasing industrial exports, the growth rate is sluggish and unemployment remains high. The government aims to reform the labor market and unemployment system, but some argue that reducing salaries for low-skilled workers is necessary to compete with countries like Hungary. The EU's goal of becoming the most efficient economy by 2010 seems increasingly distant, as challenges such as global competition and inadequate structural reforms persist. |

# BBC Article (Entertainment) 222

| Original - Reference | Suspicious - Reference | Reference - Original & Suspicious | Other - Reference |
|---|---|---|---|
| More than 1,000 people turned out on Sunday to pay tribute to late US TV presenter Johnny Carson in the Nebraska town where he grew up. Carson, who hosted the Tonight Show for 30 years, died on 23 January from respiratory disease emphysema. He lived in Norfolk, Nebraska from the age of eight until he joined the Navy, but returned regularly and donated $5m (£2.7m) to local causes. Old school friends were among the crowd at the school's Johnny Carson Theater. Carson, who was one of the best-loved TV personalities in the US, asked not to have a public memorial in Los Angeles, where he lived in later life. He began his showbusiness career in Norfolk, performing magic under the name The Great Carsoni from the age of 14. His donations included $600,000 to Norfolk High School in the 1980s to build a new performing arts centre. When Carson died, President Bush led the public tributes, saying the presenter "had a profound influence on American life and entertainment". | Over a thousand individuals gathered on a Sunday in the Nebraska town where Johnny Carson grew up, to honor the memory of the renowned US television host. Carson, famed for his 30-year tenure on the Tonight Show, passed away on January 23 due to emphysema, a respiratory condition. Carson took up residence in Norfolk, Nebraska, when he was eight and lived there until his naval service, though he often revisited and contributed $5 million to community projects. Alumni from his school days were present in the audience at the Johnny Carson Theater within the school. Known as one of the most cherished figures on American television, Carson chose not to have a public memorial in Los Angeles, where he resided during his later years. He initially entered the entertainment industry in Norfolk with his magic act as "The Great Carsoni" starting at 14 years old. His philanthropic efforts included a $600,000 gift to Norfolk High School in the 1980s for a new arts center. Following Carson's death, President Bush was at the forefront, acknowledging Carson's significant impact on both American culture and the entertainment industry. | Over a thousand individuals gathered on a Sunday in the Nebraska community where Johnny Carson was raised to honor the memory of the renowned American television host. Carson, famed for his 30-year tenure on the Tonight Show, passed away on January 23 due to the respiratory illness emphysema. He resided in Norfolk, Nebraska, from when he was eight years old until his enlistment in the Navy, though he often visited and contributed $5 million to local initiatives. Attendees at the Johnny Carson Theater in his old school included childhood friends. Carson, a dearly cherished figure on American television, preferred not to have a public memorial in Los Angeles, where he spent his later years. He initiated his career in entertainment in Norfolk with magic performances as The Great Carsoni starting at 14. His philanthropic efforts included a $600,000 gift to Norfolk High School in the 1980s for the construction of a new arts center. Following Carson's death, President Bush was at the forefront of national tributes, remarking on Carson's significant impact on American culture and the entertainment industry. | On Sunday, over a thousand people gathered in Norfolk, Nebraska, to honor Johnny Carson, the beloved former host of "The Tonight Show." Carson spent his childhood in this town and later generously supported local causes, including a $5 million donation. Carson's hometown friends joined the crowd at the school's Johnny Carson Theater to pay their respects. Despite his later fame in Los Angeles, Carson requested that his memorial take place in Norfolk, where his entertainment career began at the age of 14. President Bush mourned Carson's passing, recognizing his deep impact on American culture. Notable donations made by Carson included a substantial sum towards building Norfolk High School's performing arts center. |

# BBC Article (Entertainment) 262

**Original** - Reference   **Suspicious** - Reference   **Reference** - Original & Suspicious   **Other** - Reference

U2, who have won three prestigious Grammy Awards for their hit Vertigo, are stubbornly clinging to their status as one of the biggest bands in the world. The most popular groups in the history of rock all have several things in common. The music must be inspired and appeal across generations and be distinctive, if not always groundbreaking. But such success is down to more than music. They have to be compelling performers, charismatic and intelligent enough to make good decisions and keep their feet on the ground. They also have to want it. They have to want to be the biggest band ever and not stop wanting it. The Beatles had it, the Rolling Stones still have it, REM hold onto it and Queen were it in a catsuit. And U2 have it in spades, and keep churning it out. Their new album, How To Dismantle An Atomic Bomb, comes 28 years after the schoolfriends got together in Dublin and 17 years after The Joshua Tree cemented their place on the all-time rock A-list. They may have lost some of the edginess and raw, youthful force that propelled them to the top, but they have lost none of the desire or ability to craft songs and albums. Vertigo, the first single from the new album, went straight into the UK singles chart at number one, knocking Eminem off the top spot and giving them their 26th top 10 hit. "The challenge is to be bigger and bolder and better to make records the whole world will listen to," Bono recently said. Drummer Larry Mullen Jr echoed those sentiments: "We're very competitive we want to be on the radio, have big singles. We don't want to be thought of as a veteran band." The band have done "everything in their considerable powers" to ensure they remain the biggest band in the world, according to Q magazine editor Paul Rees. "This makes them hugely determined and formidable." He added: "They are equally determined to push themselves to make music that continues to stand up. "As such, they've constantly re-invented and challenged themselves. They are, perhaps, alone as the only rock band that has got better with age." The other key ingredient was the fact they were highly organised, Mr Rees said. "They do everything in the right way." The group were born when Mullen put an appeal for bandmates on a high school notice board, attracting fellow pupils Paul Hewson (Bono, vocals), Adam Clayton (bass), David Evans (The Edge, guitar) and his brother Dick. Dick Evans soon dropped out and the four-piece were known as The Feedback and The Hype before settling on U2. By 1978, they had won a talent contest and got noticed by a manager, Paul McGuinness. "They were brilliant, but very coarse," McGuinness recently said. "In a way, they were doing exactly what they do now. Only badly." They struggled to attract record company attention, later being described as "pretty damn average" and "strange and eerie" by scouts who saw them live. They released two Ireland-only singles, which topped the national charts in 1979 and 1980, leading to a deal with Island and their debut album Boy. The stadium-filling, anthemic sound was U2's aim from the start, and their third album, War, saw them make the breakthrough on both sides of the Atlantic, going to number one in the UK and 12 in the US. Songs like Sunday Bloody Sunday and New Year's Day brought success and an image as a political and spiritual band which Bono rejected as a cliche. His stage performances which included flag-waving, speaker-climbing and drum-throwing earned him a reputation as an electric performer, and their appearance at 1985's Live Aid is widely seen as sealing their global stardom. In 1987, The Joshua Tree broke sales records and saw the band reach the height of their powers with hits including Where the Streets Have No Name, I Still Haven't Found What I'm Looking For and With Or Without You. Those songs took the band's epic, atmospheric sound to a simple, powerful and popular pinnacle. The end of the decade marked a crucial point for the band they had reached the top but still yearned for new challenges and achievements. These came in the form of explorations of different branches of rock and forays into electronic dance music, plus wildly extravagant stage shows, while still trying to retain their mass appeal. The Achtung Baby album in 1991 was followed by Zooropa, Pop and their corresponding stadium tours, which featured giant olives, flying cars, live phone calls to the White House and Bono's transformation into alter-egos The Fly and MacPhisto. He was also building a parallel reputation not always to the pleasure of his bandmates as a campaigner on issues from global debt to Aids. Before the release of How To Dismantle An Atomic Bomb, they had sold 125 million albums around the world. But they still want more.

U2, triumphant in securing three distinguished Grammy Awards for their successful track "Vertigo", are tenaciously holding onto their title as a dominant global musical force. The most iconic rock groups share several attributes: their music is imaginative, resonates across ages, and is unique, sometimes even pioneering. Yet achievement goes beyond mere musicianship. These artists captivate with their stage presence, exhibit charisma, and are adept at making savvy choices, remaining humble. Their ambition is key—they relentlessly pursue the title of the greatest band with unwavering fervor. The Beatles embodied this, the Rolling Stones retain it, REM clings to it, and Queen personified it in spandex. U2 possesses this quality abundantly and continues to produce it. Their album "How To Dismantle An Atomic Bomb" is released 28 years post their formation in Dublin and 17 years subsequent to "The Joshua Tree" securing their legendary status. Their edge and youthful dynamism may have softened, but their songwriting prowess remains intact. "Vertigo," the lead single of their latest album, debuted atop the UK charts, besting Eminem, marking their 26th top 10 entry. Bono has emphasized their goal to push boundaries and appeal universally, and Larry Mullen Jr reflected the band's competitive spirit—they aspire to airplay and hit songs, refusing to be labeled as simply enduring. Q magazine's editor Paul Rees commends their intense determination and capability, suggesting they are unique in improving with age, also noting their exceptional organizational skills. The genesis of U2 was when Mullen advertised for bandmates on a school noticeboard, drawing in Bono, Adam Clayton, and David Evans (The Edge), along with Evans' brother who soon left the then-named Feedback and The Hype, eventually settling on U2. By 1978, their talent competition win attracted manager Paul McGuinness, and despite an initial struggle and limited record label interest, they achieved national chart success in Ireland with early singles. Their distinctive sound was evident from the get-go, and their third record, "War," earned them a major international presence. Tracks like "Sunday Bloody Sunday" and "New Year's Day" positioned them as a band inclined towards political and spiritual themes, which Bono refuted as oversimplified. Their energized live performances, notably at Live Aid in 1985, cemented their international fame. With hits from "The Joshua Tree" in 198

U2, the recipients of three esteemed Grammy Awards for their song "Vertigo," tenaciously maintain their position as one of the world's most prominent bands. All legendary rock groups share certain traits: their music must be innovative, resonate with multiple generations, and be unique, even if not always revolutionary. However, their triumph extends beyond the music. They must be engaging on stage, possess charisma, and have the intelligence to make wise choices while staying humble. They must also possess a relentless ambition to be the greatest band ever. This drive was evident in The Beatles, persists in the Rolling Stones, is maintained by REM, and was embodied by Queen in a catsuit. U2 has this quality in abundance and continues to produce it. Their latest album, "How To Dismantle An Atomic Bomb," arrives 28 years after the band formed in Dublin as school friends and 17 years following "The Joshua Tree," which secured their spot among rock's elite. While they may have shed some of their initial edginess and raw energy, their passion and skill for creating music remain undiminished. "Vertigo," the lead single from the new album, debuted at number one in the UK singles chart, giving them their 26th top 10 hit and displacing Eminem from the top. Bono has expressed the ambition to create music that captivates the entire world, while drummer Larry Mullen Jr has shared the band's competitive nature, emphasizing their desire for radio play and hit singles without being labeled as a veteran act. According to Q magazine's editor Paul Rees, U2 has exerted all their considerable effort to stay at the pinnacle of the music world, making them a determined and formidable force. He notes that they are committed to pushing their musical boundaries and have uniquely improved with age. Rees also credits their success to their high level of organization. The band's origins trace back to a high school notice board where Mullen sought bandmates, leading to the formation of U2 with Paul Hewson (Bono), Adam Clayton, David Evans (The Edge), and his brother Dick, who later left the group. They went through names like The Feedback and The Hype before settling on U2. By 1978, they had caught the attention of manager Paul McGuinness after winning a talent contest. Despite initial struggles to gain record label interest and mixed reviews from scouts, they released two singles that topped the Irish charts in 1979

Despite winning three Grammys for their hit single "Vertigo," U2 remains determined to maintain their status as one of the biggest bands in the world. Legendary rock groups share common traits: compelling music that transcends generations, distinctiveness, exceptional live performances, charisma, intelligence, and an unwavering desire for greatness. U2 embodies these qualities, evidenced by their continued success 28 years after their formation. Their new album reflects their enduring passion for crafting unforgettable songs. The band's success is not solely due to their music. They are driven by a need to be the best, to create records that resonate with global audiences. Drummer Larry Mullen Jr. emphasizes their competitive spirit and desire for success. Editor Paul Rees highlights their meticulous organization and dedication to pushing musical boundaries. U2's journey began with a high school band featuring Bono, The Edge, Adam Clayton, and Larry Mullen Jr. Their persistence eventually led to a record deal and the release of their debut album, "Boy." Their breakthrough came with "War," solidifying their status on both sides of the Atlantic. Bono's captivating stage presence and their political and spiritual messages have contributed to their global fame. Live Aid cemented their stardom, and "The Joshua Tree" showcased their epic and popular sound. However, the band yearned for more, leading them to experiment with different genres and extravagant stage shows. Despite selling 125 million albums worldwide, U2 remains hungry for continued success. Their enduring passion and drive have been instrumental in their longevity and global dominance.

# BBC Article (Politics) 333

| Original - Reference | Suspicious - Reference | Reference - Original & Suspicious | Other - Reference |
|---|---|---|---|
| Ex-chat show host Robert Kilroy-Silk is to contest the Derbyshire seat of Erewash at the next general election. Labour's Elizabeth Blackman won the seat in 1997 and has a 6,932 majority. She says she will fight on her record "as a hard-working constituency MP". Mr Kilroy-Silk announced his plans a day after launching his new party, Veritas, the Latin for truth. The East Midlands MEP, who quit the UK Independence Party, wants his new group to "change the face" of UK politics. His choice of election constituency quashes speculation that he would stand against Defence Secretary Geoff Hoon in Ashfield, Nottinghamshire. UKIP won 31% of the vote in Erewash in last June's European elections with Mr Kilroy-Silk among their candidates for the region. Until 1997, Erewash had been held by the Tories since 1970. Ms Blackman said she was proud of the government's achievements in the area. She declined to give her view of Mr Kilroy-Silk at this point. On Thursday, he told a London news conference that Veritas would avoid the old parties' "lies and spin". He said "our country" was being "stolen from us" by mass immigration and promised a "firm but fair" policy on immigration. Veritas says it hopes to contest most seats at the forthcoming general election but plans to announce detailed policies on crime, tax, pensions, health and defence over the next few weeks. UKIP leader Roger Knapman says he is glad to see the back of Mr Kilroy-Silk. Labour campaign spokesman Fraser Kemp said Veritas was joining "an already crowded field on the right of British politics". Mr Kilroy-Silk was joined in the new venture by one of UKIP's two London Assembly members, Damien Hockney, who is now Veritas' deputy leader. UKIP's chairman Petrina Holdsworth has said the group will just be a parody of the party the men have left. Mr Kilroy-Silk quit UKIP last week after months of tension as he vied unsuccessfully for the leadership of that party. He said he was ashamed to be a member of a UKIP whose leadership had "gone AWOL" after the great opportunity offered by its third place at last June's European elections. UKIP's leader, Roger Knapman, has said he is glad to see the back of Mr Kilroy-Silk. "He has remarkable ability to influence people but, sadly, after the [European] election it became clear that he was more interested in the Robert Kilroy-Silk Party than the UK Independence Party so it was nice knowing him, now 'goodbye'," he said. UKIP officials also argue Mr Kilroy-Silk has not been "straightforward" in attacking the party he once wanted to lead. | Former television presenter Robert Kilroy-Silk has announced his intention to run for the parliamentary seat of Erewash in Derbyshire at the upcoming general election. The seat is currently held by Labour's Elizabeth Blackman, who secured it in 1997 with a majority of 6,932 votes. Blackman has stated she will campaign based on her reputation as a diligent local MP. Kilroy-Silk revealed his intention to run shortly after introducing his newly-formed political party, Veritas, which means "truth" in Latin. Having recently left the UK Independence Party (UKIP), Kilroy-Silk is now an MEP for the East Midlands and aims to bring transformative change to British politics with his new party. He has ended rumors that he might challenge Defence Secretary Geoff Hoon in the Ashfield constituency of Nottinghamshire. In Erewash, UKIP garnered 31% of the vote in the European elections of the previous June, with Kilroy-Silk as one of their candidates. The Conservative Party held the Erewash seat from 1970 until 1997. Blackman has expressed her confidence in the government's work in the local area but refrained from commenting on Kilroy-Silk's candidacy. During a press briefing in London, Kilroy-Silk criticized the deceit and manipulations of established political parties and expressed concerns over mass immigration, promising a balanced immigration policy. Veritas is planning to run candidates in most constituencies and will reveal its positions on issues such as crime, taxation, pensions, health care, and defense in the forthcoming weeks. Roger Knapman, the leader of UKIP, has expressed relief at Kilroy-Silk's departure from the party. Labour's campaign spokesperson Fraser Kemp remarked that Veritas is entering an already competitive right-wing political scene. The new party's deputy leader is Damien Hockney, a former UKIP member of the London Assembly. UKIP's chairperson, Petrina Holdsworth, characterized Veritas as a mere imitation of the party the two men have abandoned. Kilroy-Silk resigned from UKIP due to disagreements and disappointment with the party's leadership, which he felt was absent following UKIP's notable performance in the European elections. Knapman acknowledges Kilroy-Silk's persuasive talents but criticizes his focus on personal ambitions after the elections, leading to his departure. UKIP officials also claim that Kilroy-Silk has been deceitful in his public critique of the party he once attempted to lead. | Former television presenter Robert Kilroy-Silk is set to run for the parliamentary seat in Erewash, Derbyshire, in the upcoming general election. The seat is currently held by Labour's Elizabeth Blackman, who secured it in 1997 with a majority of 6,932 votes. Blackman has expressed her intention to campaign based on her reputation as a diligent local MP. Kilroy-Silk revealed his candidacy shortly after introducing his new political party, Veritas, which means "truth" in Latin. As a Member of the European Parliament for the East Midlands and a former member of the UK Independence Party (UKIP), he aims for his new party to significantly impact British politics. His decision to run in Erewash ends speculation that he might challenge Defence Secretary Geoff Hoon in the Ashfield constituency of Nottinghamshire. In the European elections of the previous June, UKIP garnered 31% of the vote in Erewash, with Kilroy-Silk as one of their candidates. The Conservatives had previously held Erewash from 1970 until 1997. Blackman has expressed pride in the government's local achievements but has refrained from commenting on Kilroy-Silk's candidacy. On Thursday, Kilroy-Silk criticized the established political parties for their dishonesty at a press conference in London, claiming that the nation was being undermined by extensive immigration. He promised a balanced immigration policy. Veritas plans to run candidates in most constituencies in the next general election and will unveil its policies on crime, taxation, pensions, healthcare, and defense in the coming weeks. Roger Knapman, the leader of UKIP, expressed relief at Kilroy-Silk's departure from the party. Labour campaign spokesperson Fraser Kemp remarked that Veritas is entering an already competitive right-wing political landscape. Kilroy-Silk has been supported in his new endeavor by Damien Hockney, a former UKIP London Assembly member who now serves as the deputy leader of Veritas. Petrina Holdsworth, UKIP's chairperson, has dismissed Veritas as a mere imitation of the party they have abandoned. Kilroy-Silk left UKIP after unsuccessfully attempting to become its leader, criticizing the party's leadership for being absent following their strong performance in the European elections. Knapman has acknowledged Kilroy-Silk's persuasive talents but noted that post-election, it became apparent that Kilroy-Silk was more focused on his own interests than those of UKIP. UKIP officials have | Robert Kilroy-Silk, the former talk show host, will seek election as a representative for Erewash. The incumbent, Labour's Elizabeth Blackman, will defend her seat, which she has held since 1997. Kilroy-Silk formed Veritas, a new political party, and introduced it before announcing his candidacy. The East Midlands MEP resigned from UKIP to lead Veritas's mission to transform British politics. He dismissed suggestions that he might challenge Defense Secretary Geoff Hoon in Nottinghamshire. During the European elections in June, UKIP secured 31% of votes in Erewash, where Kilroy-Silk was among the candidates. Since 1970, Erewash had been under Conservative rule but was won by Ms. Blackman in 1997. Ms. Blackman expressed pride in her government's achievements and declined to comment on Kilroy-Silk's candidacy. Veritas intends to distance itself from "lies and spin" employed by established parties. Kilroy-Silk highlighted concerns about mass immigration and pledged to implement a "firm but fair" immigration policy. Veritas will contest a majority of seats in the upcoming election while unveiling detailed policies on crime, healthcare, and defense later. UKIP's leader, Roger Knapman, expressed relief at Kilroy-Silk's departure. Damien Hockney, one of UKIP's London Assembly members, has joined Kilroy-Silk's new venture as deputy leader. UKIP officials believe Veritas will be a mere imitation of their organization. Kilroy-Silk resigned from UKIP after facing tensions during an unsuccessful bid for party leadership. He criticized UKIP's leadership for not capitalizing on their success in the European elections and accused them of abandoning their supporters. Knapman welcomed Kilroy-Silk's resignation, stating that he had swayed many supporters but prioritized his own interests. UKIP officials have also expressed concerns over Kilroy-Silk's criticism of the party he once sought to lead. |

# BBC Article (Politics) 351

| Original - Reference | Suspicious - Reference | Reference - Original & Suspicious | Other - Reference |
|---|---|---|---|
| Tony Blair lied when he took the UK to war so has no qualms about lying in the election campaign, say the Tories. Tory co-chairman Liam Fox was speaking after Mr Blair told Labour members the Tories offered a "hard right agenda". Dr Fox told BBC Radio: "If you are willing to lie about the reasons for going to war, I guess you are going to lie about anything at all." He would not discuss reports the party repaid £500,000 to Lord Ashcroft after he predicted an election defeat. The prime minister ratcheted up Labour's pre-election campaigning at the weekend with a helicopter tour of the country and his speech at the party's spring conference. He insisted he did not know the poll date, but it is widely expected to be 5 May. In what was seen as a highly personal speech in Gateshead on Sunday, Mr Blair said: "I have the same passion and hunger as when I first walked through the door of 10 Downing Street." He described his relationship with the public as starting euphoric, then struggling to live up to the expectations, and reaching the point of raised voices and "throwing crockery". He warned his supporters against complacency, saying: "It's a fight for the future of our country, it's a fight that for Britain and the people of Britain we have to win." Mr Blair said that whether the public chose Michael Howard or Mr Kennedy, it would result in "a Tory government not a Labour government and a country that goes back and does not move forward". Dr Fox accused Mr Blair and other Cabinet ministers of telling lies about their opponents' policies and then attacking the lies. "What we learned at the weekend is what Labour tactics are going to be and it's going to be fear and smear," he told BBC News. The Tory co-chairman attacked Labour's six new pledges as "vacuous" and said Mr Blair was very worried voters would take revenge for his failure to deliver. Dr Fox refused to discuss weekend newspaper reports that the party had repaid £500,000 to former Tory Treasurer Lord Ashcroft after he said the party could not win the election. "We repay loans when they are due but do not comment to individual financial matters," he said, insisting he enjoyed a "warm and constructive" relationship to Lord Ashcroft. Meanwhile Lib Dem leader Charles Kennedy is expected to attack Mr Blair's words as he begins a nationwide tour on Monday. Mr Kennedy is accelerating Lib Dem election preparations this week as he visits Manchester, Liverpool, Leicester, Somerset, Basingstoke, Shrewsbury, Dorset and Torbay. He said: "This is three-party politics. In the northern cities, the contest is between Labour and the Liberal Democrats. "In southern and rural seats especially in the South West the principal contenders are the Liberal Democrats and the Conservatives, who are out of the running in Scotland and Wales." The Lib Dems accuse Mr Blair of making a "touchy-feely" speech to Labour delegates which will not help him regain public trust. | The Conservative party, with co-chairman Liam Fox at the helm, has accused former Prime Minister Tony Blair of dishonesty, both in his decision to go to war and throughout the election campaign. Fox commented on Blair's assertion that the Tories hold a "hard right agenda," by suggesting that someone who misrepresents reasons for war can't be trusted in other matters either. Despite avoiding a direct response to rumors of the Tories returning £500,000 to Lord Ashcroft post an unfavorable election forecast, Fox criticized Labour's campaign tactics, labeling them as fearmongering and slanderous, and dismissed Labour's new commitments as meaningless. He expressed concern over the electorate's potential backlash due to Blair's unfulfilled promises. Blair boosted Labour's pre-election efforts with a nationwide helicopter tour and at the spring conference, he passionately reflected on his political journey and cautioned his followers against complacency. Blair contended that a victory for either Michael Howard or Mr. Kennedy would essentially mean a step backward under a Conservative government. Meanwhile, Liberal Democrat leader Charles Kennedy, who has embarked on his own tour, asserted that the real competition in northern urban areas is between Labour and the Lib Dems while in the south, it is between the Conservatives and the Lib Dems. Kennedy and his party dismissed Blair's emotional speech to his supporters as ineffective in rebuilding public trust. | The Conservatives claim that Tony Blair, having misled the country into war, would not hesitate to deceive the public during the election campaign. Liam Fox, the Conservative co-chairman, commented on Blair's assertion that the Conservatives have a "hard right agenda" by saying to BBC Radio that someone who can lie about war can lie about anything. Fox declined to address reports that the Conservatives returned £500,000 to Lord Ashcroft after he anticipated an election loss. Over the weekend, Blair intensified Labour's pre-election efforts with a helicopter tour and a speech at the party's spring conference. Although he claimed ignorance of the exact election date, it is widely anticipated to be on May 5th. In Gateshead, Blair delivered a personal speech, expressing his undiminished zeal since his first day at 10 Downing Street. He recounted the highs and lows of his relationship with the public, framing the election as a crucial battle for Britain's future. Blair cautioned that choosing either Michael Howard or Mr. Kennedy would lead to a Conservative government and a step backward for the country. Dr. Fox criticized Blair and his ministers for allegedly lying about the opposition's policies and then attacking those fabrications, labeling Labour's strategy as one of fear and smear. He dismissed Labour's six new pledges as empty and suggested that Blair feared electoral retribution for his unfulfilled promises. Fox also declined to comment on the repayment of £500,000 to Lord Ashcroft, stating that the party repays loans when due but does not discuss individual financial matters, and he described his relationship with Lord Ashcroft as positive. Meanwhile, Liberal Democrat leader Charles Kennedy is set to criticize Blair's approach as he embarks on a national tour. Kennedy is stepping up the Lib Dems' election campaign with visits to various cities, emphasizing the three-party nature of politics and the regional contests between different parties. The Lib Dems criticize Blair's speech to Labour delegates as insincere and doubt it will restore public trust. | The Conservative Party alleges that Prime Minister Tony Blair's untruths about the reasons for going to war indicate his willingness to deceive in the upcoming election campaign. Tory co-chairman Liam Fox claims that Blair's earlier deception suggests a lack of integrity or qualms about lying. However, Fox declined to address rumors of a £500,000 repayment to Lord Ashcroft following Ashcroft's prediction of an election loss. Over the weekend, Blair intensified Labour's pre-election efforts with a helicopter tour and a speech at the party's conference. Despite denying knowledge of the election date, it is widely believed to be May 5th. In an emotionally charged speech, Blair urged supporters against complacency, emphasizing the critical nature of the fight for Britain's future. Dr. Fox accused Blair and his cabinet of fabricating lies about opponents' policies and then criticizing those falsehoods. He characterized Labour's tactics as fear-mongering and smear campaigns. He also criticized Labour's six new pledges as empty and expressed Blair's concern over voters' retribution for his perceived failures. Meanwhile, Liberal Democrat leader Charles Kennedy is expected to criticize Blair's words as he embarks on a nationwide tour. Kennedy is ramping up Lib Dem election preparations, visiting various cities and regions. He highlighted the varied nature of the electoral contest, noting that the Liberal Democrats face different primary competitors in different areas. The Lib Dems view Blair's speech to Labour delegates as self-serving and ineffective in regaining public trust. |

# BBC Article (Sports) 440

| Original - Reference | Suspicious - Reference | Reference - Original & Suspicious | Other - Reference |
|---|---|---|---|
| Veteran Spaniard Conchita Martinez came from a set down to beat Italian Roberta Vinci at the Qatar Open in Doha. The 1994 Wimbledon champion won 5-7 6-0 6-2 to earn a second round meeting with French Open champion Anastasia Myskina. Fifth seed Patty Schnyder also had a battle as she needed three sets to beat China's Na Li 7-5 3-6 7-5. Slovakian Daniela Hantuchova beat Bulgarian Magdaleena Maleeva 4-6 6-4 6-3 to set up a second round clash with Russian Elena Bovina. The veteran Martinez found herself in trouble early on against Vinci with the Italian clinching the set thanks to breaks in the third and 11th games. But Vinci's game fell to pieces after that and Martinez swept her aside with some crisp cross-court returns and deft volleys. In the day's other matches, Japan's Ai Sugiyama defeated Australian Samantha Stosur 6-2 6-3 while Australian Nicole Pratt beat Tunisian Selima Sfar 7-5 6-2 and will next face compatriot Alicia Molik. | Experienced Spanish player Conchita Martinez rebounded after losing the initial set to defeat Italy's Roberta Vinci at the Qatar Open in Doha. The winner of the 1994 Wimbledon title triumphed with scores of 5-7, 6-0, 6-2, setting up a second-round encounter with French Open winner Anastasia Myskina. Meanwhile, fifth-seeded Patty Schnyder was pushed to her limits, as she took three sets to overcome China's Na Li with a scoreline of 7-5, 3-6, 7-5. Slovakia's Daniela Hantuchova overcame Bulgaria's Magdaleena Maleeva 4-6, 6-4, 6-3, paving the way for her next match against Russia's Elena Bovina. Martinez initially struggled against Vinci, who secured the first set by breaking serve in the third and eleventh games. However, Vinci's performance deteriorated thereafter, and Martinez dominated by executing sharp cross-court shots and skillful net play. Among other matches of the day, Ai Sugiyama of Japan triumphed over Samantha Stosur of Australia with a clear 6-2, 6-3 victory, while Australia's Nicole Pratt won against Tunisia's Selima Sfar 7-5, 6-2, and is slated to play against fellow Australian Alicia Molik next. | Experienced Spanish player Conchita Martinez recovered after losing the first set to defeat Italy's Roberta Vinci at the Qatar Open in Doha. The winner of Wimbledon in 1994 turned the match around with a 5-7, 6-0, 6-2 victory, setting up a second-round encounter with the French Open winner Anastasia Myskina. Meanwhile, the fifth-seeded Patty Schnyder was pushed to three sets before overcoming China's Na Li with scores of 7-5, 3-6, 7-5. From Slovakia, Daniela Hantuchova overcame Bulgaria's Magdaleena Maleeva 4-6, 6-4, 6-3, preparing for her next match against Russia's Elena Bovina. Martinez initially struggled against Vinci, who secured the first set by breaking serve in the third and eleventh games. However, Vinci's performance deteriorated thereafter, and Martinez dominated with sharp cross-court shots and skillful net play. In other matches of the day, Japan's Ai Sugiyama triumphed over Australia's Samantha Stosur with a score of 6-2, 6-3, and Australia's Nicole Pratt won against Tunisia's Selima Sfar 7-5, 6-2, setting up a match with her fellow Australian Alicia Molik. | Conchita Martinez, a seasoned Spaniard, rallied from a set down to triumph over Roberta Vinci of Italy at the Qatar Open in Doha. Martinez, the 1994 Wimbledon victor, prevailed with a score of 5-7, 6-0, 6-2, earning her a second-round encounter against Anastasia Myskina, the French Open champion. Patty Schnyder, seeded fifth, faced a tough battle as she needed three sets to overcome China's Na Li, eventually winning 7-5, 3-6, 7-5. Daniela Hantuchova of Slovakia defeated Bulgarian Magdaleena Maleeva 4-6, 6-4, 6-3, setting up a second-round matchup with Russian Elena Bovina. In the early stages of her match against Vinci, veteran Martinez faced adversity, with the Italian seizing the set through breaks in the third and eleventh games. However, Vinci's performance deteriorated drastically thereafter, as Martinez overpowered her with precise cross-court returns and skillful volleys. Other matches of the day saw Ai Sugiyama of Japan defeat Samantha Stosur of Australia 6-2, 6-3. Nicole Pratt of Australia emerged victorious against Selima Sfar of Tunisia 7-5, 6-2, and will face her fellow Australian, Alicia Molik, in the next round. |

# BBC Article (Sports) 211

| Original - Reference | Suspicious - Reference | Reference - Original & Suspicious | Other - Reference |
|---|---|---|---|
| Arsenal's Brazilian midfielder Edu has hit out at the club for stalling over offering him a new contract. Edu's deal expires next summer and he has been linked with Spanish trio Real Madrid, Barcelona and Valencia. He told BBC Sport: "I'm not sure if I want to stay or not because the club have let the situation go on this far. "If they had really wanted to sign they should have come up with an offer six months before indicating they wanted to sign me and that's made me think." Edu's brother and representative Amadeo Fensao has previously said that Arsenal's current offer to the midfielder was well short of what he was seeking. And Edu, 26, added: "My brother is due to come to London on Thursday. "There is a meeting planned for 6 or 7 January to sort it out with Arsenal. "Now I have a choice to stay or go. I want to sort it out as soon as possible, that's in the best interests of both the club and myself. "I'm going to make my decision after the meeting later this week." Edu is now able to begin negotiations with other clubs because Fifa regulations allow players to start talks six months before their contracts expire. The midfielder, who broke in to the Brazilian national side in 2004, admitted he had been flattered to have been linked with the three Spanish giants. Edu said: "I've just heard stories from the news that the Madrid president Florentino Perez, the Valencia people, as well as Barcelona are interested. "That's nice, but I've never talked to them, so I can't say they want me sign 100%." Last month Wenger said he was hopeful Edu would sign a new deal and played down suggestions that the lure of a club like Real Madrid would be too strong for Edu. Edu added that he had been encouraged by Wenger's support for him. "I still have a good relationship with Arsene Wenger he's always said he wants me to sign." | Arsenal's Brazilian midfielder, Edu, has expressed frustration with the club for delaying a new contract offer. With his current contract expiring next summer, Edu has attracted interest from Spanish clubs Real Madrid, Barcelona and Valencia. In a discussion with BBC Sport, Edu remarked, "The club's lack of urgency makes me unsure about my desire to stay. They should have presented an offer six months earlier to show their intent to keep me, and their failure to do so has caused me to consider my options." Edu's brother and agent, Amadeo Fensao, has mentioned that the offer from Arsenal falls short of their expectations. Edu, 26 years old, stated, "My brother is coming to London on Thursday, and we've scheduled a meeting with Arsenal on the 6th or 7th of January to resolve this. Im faced with the decision to either stay or leave, and I aim to settle this matter promptly for the benefit of everyone involved. My decision will come after this upcoming meeting." Now eligible to negotiate with other clubs due to FIFA regulations permitting players to enter talks six months before their contract's end, Edu acknowledged the honor of being associated with such prestigious Spanish teams. Despite the rumors, he confirmed, "I haven't spoken to them directly, so I can't confirm their interest in signing me." Last month, Arsenal manager Arsene Wenger expressed optimism about Edu re-signing and downplayed the appeal that a team like Real Madrid might have. Edu noted Wenger's continued support and his stated desire for Edu to re-sign with the club. | Arsenal's Brazilian midfielder Edu has expressed frustration with the club for delaying a new contract offer. With his current contract set to expire next summer, Edu has been linked with a move to Spanish clubs Real Madrid, Barcelona, and Valencia. Speaking to BBC Sport, Edu voiced uncertainty about his future at Arsenal, stating, "I'm not sure if I want to stay or not because the club have let the situation go on this far. If they had really wanted to sign they should have come up with an offer six months before indicating they wanted to sign me and that's made me think." Edu's brother and agent, Amadeo Fensao, has mentioned that the offer from Arsenal falls short of their expectations. Edu, who is 26 years old, mentioned that his brother is scheduled to arrive in London for a meeting with Arsenal on either January 6 or 7 to resolve the matter. "Now I have a choice to stay or go. I want to sort it out as soon as possible, that's in the best interests of both the club and myself. I'm going to make my decision after the meeting later this week," Edu stated. Under Fifa rules, Edu is now eligible to start discussions with other clubs, as he is within six months of his contract's end. Having made his way into the Brazilian national team in 2004, Edu acknowledged that the interest from the Spanish powerhouses was flattering, but clarified that he had not engaged in talks with them, saying, "That's nice, but I've never talked to them, so I can't say they want me sign 100%." Arsenal manager Arsene Wenger had previously expressed optimism that Edu would renew his contract and downplayed the idea that a club like Real Madrid could sway Edu's decision. Edu appreciated Wenger's support, noting, "I still have a good relationship with Arsene Wenger he's always said he wants me to sign." | After stalling over a new contract, Arsenal's Edu has expressed his dissatisfaction with the club. Despite his deal expiring next summer, Edu remains indecisive about his future as negotiations have dragged on. His agent's previous announcement indicates that Arsenal's offer falls short of his expectations. With his contract expiring within six months, Edu is free to explore offers from other clubs. The Brazilian international acknowledges the flattering interest from Real Madrid, Barcelona, and Valencia but emphasizes that no concrete discussions have taken place. Edu's decision will come after a scheduled meeting with Arsenal later this week, weighing his options between remaining or seeking opportunities elsewhere. Despite Wenger's public support, Edu highlights the missed opportunity for a timely contract offer that could have swayed his loyalty. |

# BBC Article (Technology) 339

| Original - Reference | Suspicious - Reference | Reference - Original & Suspicious | Other - Reference |
|---|---|---|---|
| People using wireless net hotspots will soon be able to make free phone calls as well as surf the net. Wireless provider Broadreach and net telephony firm Skype are rolling out a service at 350 hotspots around the UK this week. Users will need a Skype account downloadable for free and they will then be able to make net calls via wi-fi without paying for net access. Skype allows people to make free PC-based calls to other Skype users. Users of the system can also make calls to landlines and mobiles for a fee. The system is gaining in popularity and now has 28 million users around the world. Its paid service dubbed Skype Out has so far attracted 940,000 users. It plans to add more paid services with forthcoming launches of video conferencing, voice mail and Skype In, a service which would allow users to receive phone calls from landlines and mobiles. London-based software developer Connectotel has unveiled software that will expand the SMS functions of Skype, allowing users to send text messages to mobile phones from the service. Broadreach Networks has around two million users and hotspots in places such as Virgin Megastores, the Travelodge chain of hotels and all London's major rail terminals. The company is due to launch wi-fi on Virgin Trains later in the year. "Skype's success at spreading the world about internet telephony is well-known and we are delighted to be offering free access to Skype users in our hotspots," commented Broadreach chief executive Magnus McEwen-King. | Individuals utilizing Wi-Fi hotspots will soon have the option to place phone calls at no cost in addition to browsing the internet. Broadreach, a wireless service provider, in collaboration with internet phone company Skype, is introducing a service to 350 hotspots throughout the UK this week. To use this service, customers must have a free Skype account, and once they have it, they can make internet calls via Wi-Fi without incurring internet access charges. Skype enables users to call other Skype users for free from their computers. The platform also offers the ability to call traditional landlines and mobile phones for a certain price. The service is becoming more widespread, currently boasting 28 million users globally. Its premium service, known as Skype Out, has gained nearly 940,000 customers. Skype is planning to expand its paid offerings with new features such as video conferencing, voicemail, and Skype In, which will enable users to receive calls from landlines and mobile phones. Connectotel, a software company based in London, has developed software that broadens Skype's SMS capabilities, permitting users to send text messages to mobile phones through Skype. Broadreach Networks, with an approximate user base of two million, operates hotspots in locations that include Virgin Megastores, Travelodge hotel chain, and major railway stations in London. They are set to introduce Wi-Fi services on Virgin Trains later in the year. "Skype's reputation for popularizing internet telephony is widely recognized, and we're excited to provide Skype users with complimentary access at our hotspots," stated Broadreach CEO Magnus McEwen-King. | Individuals who frequent wi-fi hotspots will soon have the option to make phone calls at no cost in addition to browsing the internet. Broadreach, a wireless service provider, in collaboration with internet phone company Skype, is introducing a new service across 350 hotspots throughout the UK this week. To use the service, individuals will need to create a Skype account, which can be obtained for free. Once set up, they can make internet calls via wi-fi without incurring internet usage charges. Skype enables users to call other Skype users for free from their computers. The service also offers the ability to call traditional landlines and mobile phones for a small charge. The popularity of Skype is on the rise, with a global user base of 28 million. Its premium service, known as Skype Out, has attracted 940,000 subscribers. Skype is looking to expand its paid offerings, including video conferencing, voicemail, and Skype In, which would let users receive calls from landlines and mobiles. Connectotel, a software company based in London, has developed software that enhances Skype's SMS capabilities, allowing messages to be sent to mobile phones from the platform. Broadreach Networks, with approximately two million users, operates hotspots in locations such as Virgin Megastores, Travelodge hotels, and major London railway stations. The company also plans to introduce wi-fi services on Virgin Trains later in the year. "Skype's effectiveness in promoting internet telephony is widely recognized, and we're excited to provide Skype users with free access at our hotspots," said Magnus McEwen-King, CEO of Broadreach. | Soon, people using Wi-Fi hotspots offered by Broadreach will be able to enjoy free phone calls in addition to internet access. Through a collaboration with Skype, users will only need a free Skype account to make calls over Wi-Fi without incurring net access charges. Skype lets users make free PC-based calls to other Skype users. The service has gained immense popularity, with 28 million users worldwide. For a fee, users can also make calls to landlines and mobile phones. Forthcoming paid services from Skype include video conferencing, voice mail, and Skype In, which allows users to receive calls from landlines and mobiles. Connectotel, a London-based software developer, has created software that enhances Skype's SMS capabilities, enabling users to send text messages to mobile phones. Broadreach Networks' two million users can access hotspots in locations including Virgin Megastores, Travelodge hotels, and major rail terminals in London. Later this year, Broadreach plans to launch Wi-Fi on Virgin Trains. Broadreach's chief executive, Magnus McEwen-King, expressed his delight in offering free Skype access to their hotspot users, acknowledging Skype's prominent role in promoting internet telephony. |

# BBC Article (Technology) 198

**Original** - Reference | **Suspicious** - Reference | **Reference** - Original & Suspicious | **Other** - Reference

---

**Original - Reference:**

Analyst Bill Thompson has seen the future and it is in his son's hands. I bought my son Max a 3G phone, partly because they are so cheap and he needed a phone, and partly because I am supposed to know about the latest technology and thought I should see how they work in real life. After using it for a while I am not at all tempted to get rid of my SonyEricsson P800 smart phone. That has a relatively large screen, even if it does only have slower GPRS access to the network. I can read my e-mail, surf the web using a proper browser and write stuff using the stylus on its touch screen. Last week someone e-mailed me a document that had been compressed into a ZIP file, and I was pleasantly surprised to discover that my phone even knew how to decompress it for me. By contrast the confusing menus, complicated keyboard and truly irritating user interface of Max's 3G phone simply get in the way, and I did not see much value in the paid-for services, especially the limited web access. The videos of entertainment news, horoscopes and the latest celebrity gossip did not appeal, and I did not see how the small screen could be useful for any sort of image, never mind micro-TV. But then Max started playing, and I realised I was missing the point entirely. It is certainly not a great overall experience, but that is largely due to the poor menu system and the phone layout: the video content itself is compelling. The quality was at least as good as the video streaming from the BBC website, and the image is about the same size. Max was completely captivated, and I was intrigued to discover that I had nearly missed the next stage of the network revolution. It is easy to be dismissive of small screens, and indeed anyone of my generation, with failing eyesight and the view that 'there's never anything worth watching on TV', is hardly going to embrace these phones. But just as the World Wide Web was the "killer application" that drove internet adoption, music videos are going to drive 3G adoption. With Vodafone now pushing its own 3G service, and 3 already established in the UK, video on the phone is clearly going to become a must-have for kids forced sitting on the school bus, adults waiting outside clubs and anyone who has time to kill and a group of friends to impress. This will please the network operators, who are looking for some revenue from their expensively acquired 3G licences. But it goes deeper than that: playing music videos on a phone marks the beginning of a move away from the 'download and play' model we have all accepted for our iPods and MP3 players. After all, why should I want to carry 60GB of music and pictures around with me in my pocket when I can simply listen to anything I want, whenever I want, streamed to my phone? Oh and of course you can always use the phone to make voice calls and send texts, something which ensures that it is always in someone's pocket or handbag, available for other uses too. I have never really approved of using the Internet Protocol (IP), to do either audio or video streaming, and I think that technically it is a disaster to make phone calls over the net using "voice over IP". But I have to acknowledge that the net, at least here in the developed Western countries, is fast and reliable enough to do both. I stream radio to my computer while I work, and enjoy hearing the bizarre stations from around the world that I can find online but nowhere else. I am even playing with internet telephony, despite my reservations, and I appear on Go Digital on the World Service, streamed over the web each week. But 3G networks have been designed to do this sort of streaming, both for voice and video, which gives them an edge over net-based IP services. The 3G services aren't quite there yet, and there is a lot to be sorted out when it comes to web access and data charges. Vodafone will let you access its services on Vodafone Live! as part of your subscription cost but it makes you pay by the megabyte to download from other sites this one, for example. This will not matter to business users, but will distort the consumer market and keep people within the phone company's collection of partner sites, something that should perhaps be worrying telecoms regulator Ofcom. But we should not see these new phones simply as cut-down network terminals. If I want fast access to my e-mail I can get a 3G card for my laptop or hook up to a wireless network. The phone is a lot more, and it is as a combination of mini-TV, personal communications device and music/video player that it really works. There is certainly room in the technology ecosystem for many different sorts of devices, accessing a wide range of services over different networks. 3G phones and iPods can co-exist, at least for a while, but if I had to bet on the long term I would go for content on demand over carrying gigabytes in my pocket. Or perhaps some enterprising manufacturer will offer me both. An MP3G player, anyone? Bill Thompson is a regular commentator on the BBC World Service programme Go Digital.

---

**Suspicious - Reference:**

Expert Bill Thompson has foreseen the role his son will play in shaping technological progress. Motivated by the affordability and his son's need for a mobile phone, as well as his own professional duty to stay abreast of new tech, Thompson purchased a 3G phone for his son Max. He remained loyal to his SonyEricsson P800, appreciating its ample screen and its capabilities despite its slower GPRS connection. His P800's functionalities like email browsing, website surfing through a standard browser, and writing with a stylus on the touch screen were satisfactory for him, even handling compressed files effortlessly. While Max's 3G phone had an off-putting interface and unimpressive paid services with restricted online browsing, Thompson wasn't initially convinced of its merits. The small screen seemed impractical for viewing images or micro-TV, and the included videos of entertainment appeared frivolous to him. However, observing Max's engagement with the phone, Thompson admitted to nearly overlooking a significant technological shift. Despite flaws in design and navigation, the video content proved to be engaging with quality comparable to other streaming platforms. As 3G services become more pervasive, driven by music videos akin to the role of the World Wide Web for the internet, network providers will benefit from this new demand that appeals to various demographics. Thompson himself acknowledges the strength of current networks to support streaming and internet calling services, despite his initial skepticism about their efficacy for professional uses like Internet Protocol (IP). 3G networks are particularly optimized for such streaming, granting them an advantage over conventional IP services. While there are issues to address regarding web access and data costs, Thompson recognizes that these devices are not just network tools but multifunctional gadgets serving as mini-TVs, communication tools, and media players. In the diverse tech landscape, Thompson predicts that the convenience of accessing content on demand will eventually outweigh the need to physically carry around large data storage, although the exact future of these technologies remains open. Bill Thompson regularly shares his insights on the BBC World Service program "Go Digital."

---

**Reference - Original & Suspicious:**

Expert Bill Thompson has glimpsed what the future holds, and it seems to be shaped by his son's experiences. I purchased a 3G phone for my son Max because of its affordability and his need for a phone, and also because, as someone expected to be knowledgeable about cutting-edge tech, I wanted to understand its practical use. Despite having used it, I'm not inclined to switch from my SonyEricsson P800 smartphone. It boasts a sizeable screen, and although it's limited to slower GPRS for network access, it allows me to check emails, browse the internet with a standard browser, and write using the stylus on its touchscreen. I was recently sent a ZIP file via email and was pleasantly surprised that my phone could unzip it. In contrast, Max's 3G phone's perplexing menus, complex keyboard, and exasperating interface are obstructions, and I found little value in the chargeable services, particularly the restricted web access. The video offerings like entertainment news, horoscopes, and celebrity gossip didn't interest me, and I doubted the small screen's utility for images or so-called micro-TV. However, watching Max engage with the phone made me realize I had completely overlooked its essence. The phone's design and menu system might not offer a great experience, but the video content itself is quite engaging. The video quality was comparable to what's streamed on the BBC website, with a similar-sized image. Max was utterly absorbed, and it dawned on me that I had almost missed a significant shift in network technology. It's easy for someone from my generation, who might have deteriorating vision and a cynical view of television content, to dismiss these small screens. Yet, just as the World Wide Web was the driving force behind internet adoption, music videos might be the catalyst for 3G uptake. With Vodafone promoting its 3G service and 3 already a player in the UK market, video on phones is set to become a staple for youngsters on school buses, adults waiting outside venues, and anyone looking to kill time and impress peers. This trend is a boon for network operators who have invested heavily in 3G licenses. This shift also signifies a move away from the 'download and play' approach we've grown accustomed to with devices like iPods and MP3 players. Why carry around 60GB of music and photos when I can stream anything I want, anytime, directly to my phone? And of course, the

---

**Other - Reference:**

Analyst Bill Thompson predicts that 3G phones, far from replacing full-featured smartphones, will instead revolutionize communication and entertainment through compelling video content. While the user interface of these phones may seem clunky, children are drawn to the videos they offer, showcasing the potential for music videos to drive wider adoption of 3G. This content will likely be accessed through streaming rather than downloads, ensuring constant availability. Despite some technical issues and tariffs, the combination of mini-TV, communication, and music/video player features makes 3G phones more than just network terminals. They complement existing devices by providing on-demand content rather than large storage capacities. Thompson believes streaming over 3G networks offers advantages over Internet Protocol (IP) services due to their design for both voice and video. Nonetheless, there are challenges to overcome, such as internet access and data charges. As the industry evolves, Thompson anticipates a future where content on demand prevails over portable storage, but he also acknowledges the potential for hybrid devices that combine the best of both worlds.